\documentclass[10pt]{IEEEtran}
\IEEEoverridecommandlockouts
\usepackage[numbers]{natbib}
\usepackage{amsmath,amssymb,amsfonts}
\usepackage{algorithmic}
\usepackage{graphicx}
\usepackage{textcomp}
\usepackage{xcolor}
\usepackage{hyperref}
\usepackage{braket}
\usepackage{amsthm} 

\theoremstyle{plain}
\newtheorem{theorem}{Theorem}
\newtheorem{corollary}{Corollary}
\newtheorem{proposition}{Proposition}
\newtheorem{lemma}{Lemma}
\theoremstyle{remark}
\newtheorem{remark}{Remark}
\theoremstyle{definition}
\newtheorem{example}{Example}

\newcommand{\innerproduct}[1]{\left\langle #1 \right\rangle}
\newcommand{\expect}[1]{\mathbb{E}\left[ #1 \right]}
\newcommand{\expecttwo}[2]{\mathbb{E}_{#2}\left[ #1 \right]}
\newcommand{\E}{\mathbb{E}}
\newcommand{\rowvector}[2]{\left(\begin{matrix}
		{#1} \\ {#2}
	\end{matrix} \right)}
\newcommand{\cov}[1]{\mathrm{cov}\left( #1 \right)}
\newcommand{\tr}[1]{\mathrm{Tr} \left( #1 \right) }

\begin{document}

\title{Projection Filtering with Observed State Increments with Applications in Continuous-Time Circular Filtering\\
}

\author{\IEEEauthorblockN{Anna Kutschireiter*, Luke Rast, Jan Drugowitsch} \\
\IEEEauthorblockA{\textit{Department of Neurobiology, Harvard Medical School}\\
Boston MA, United States \\
*Corresponding author. Email: anna\_kutschireiter@hms.harvard.edu} \\}

\maketitle

\begin{abstract}
Angular path integration is the ability of a system to estimate its own heading direction from potentially noisy angular velocity (or increment) observations.
Non-probabilistic algorithms for angular path integration, which rely on a summation of these noisy increments, do not appropriately take into account the reliability of such observations, which is essential for appropriately weighing one's current heading direction estimate against incoming information.
In a probabilistic setting, angular path integration can be formulated as a continuous-time nonlinear filtering problem (circular filtering) with observed state increments.
The circular symmetry of heading direction makes this inference task inherently nonlinear, thereby precluding the use of popular inference algorithms such as Kalman filters, rendering the problem analytically inaccessible.
Here, we derive an approximate solution to circular continuous-time filtering, which integrates state increment observations while maintaining a fixed representation through both state propagation and observational updates.
Specifically, we extend the established projection-filtering method to account for observed state increments and apply this framework to the circular filtering problem.
We further propose a generative model for continuous-time angular-valued direct observations of the hidden state, which we integrate seamlessly into the projection filter.
Applying the resulting scheme to a model of probabilistic angular path integration, we derive an algorithm for circular filtering, which we term the circular Kalman filter.
Importantly, this algorithm is analytically accessible, interpretable, and outperforms an alternative filter based on a Gaussian approximation.
\end{abstract}


\vspace{0.3cm}
\small
This article has been accepted for publication in \emph{IEEE Transactions on Signal Processing}, but has not been fully edited. Content may change prior to final publication. Citation information: DOI \href{https://doi.org/10.1109/TSP.2022.3143471}{10.1109/TSP.2022.3143471}.\\
\copyright 2022 IEEE. Personal use of this material is permitted. Permission from IEEE must be obtained for all other uses, in any current or future media, including reprinting/republishing this material for advertising or promotional purposes, creating new collective works, for resale or redistribution to servers or lists, or reuse of any copyrighted component of this work in other works.
\normalsize

\section{Introduction}

A compass is an immensely useful tool for a traveler trying to find their way in a barren and featureless landscape.
Absent such a tool, the traveler must employ dead reckoning, using what they roughly know about how often and how much they have turned, and summing up those turns to maintain an internal sense of their spatial orientation.
Angular path integration, i.e.,~estimation of heading direction or orientation based on angular self-motion cues, plays an essential role in spatial navigation of humans, other animals and robots \citep{heinze_principles_2018,kreiser_neuromorphic_2018}.
Imperfect sensors make angular path integration an inherently noisy process, and inevitably lead to an accumulation of error in the heading estimate over time.
Other cues, such as those from visual landmarks, can help correct the estimate's error, despite being also noisy and ambiguous.
Importantly, properly combining path integration with these external cues requires a reliability-weighted update of the orientation estimate.
Computing with uncertainties in such a strategic way is a hallmark of dynamic Bayesian inference, and calls for a probabilistic description.
Our goal in this work is to derive a dynamic probabilistic algorithm for angular path integration.

It is well known that many organisms are able to maintain an internal compass which they update by self-motion cues.
Since the discovery of orientation-selective head-direction cells in rats \citep{taube_head-direction_1990}, and, more recently, the heading direction circuit in Drosophila \citep{seelig_neural_2015}, theoretical efforts to unravel the mechanism of angular path integration in the brain have highlighted the role of angular velocity observations of self-initiated turns \citep{skaggs_model_1995,turner-evans_angular_2017}, e.g., from proprioceptive or vestibular feedback, or from visual flow.
Current theories suggest that these biological systems implement angular path integration by neural network motifs called ring attractors \citep{xie_double-ring_2002,turner-evans_angular_2017}.
Such networks maintain a heading direction estimate through sustained neural activity, but lack the ability to simultaneously represent the estimate's certainty.
However, the question of whether such biological systems indeed only operate with single point estimates or instead perform probabilistic inference has been hampered by the lack of a \emph{probabilistic} algorithm for path integration in the brain.
The reason for this is the complex set of conditions that are required of such an algorithm:
(i) the state-space is circular,
(ii) the path integration must operate with a continuous time stream of inputs,
(iii) it must maintain a fixed representation of the underlying probability distribution, in line with the expectation that, within a certain area of computation, the brain maintains a similarly fixed representation e.g.,~in terms of a parametric \citep{ma_bayesian_2006} or sampling-based \citep{fiser_statistically_2010} representation, and
(iv)~in addition to angular velocity observations, (noisy) direct angular observations (e.g.,~visual landmarks) may be present, which need to be integrated accordingly.


One approach to designing an algorithm that satisfies these conditions is to consider it in the broader context of continuous-time filtering, which aims to continuously update a posterior distribution over a dynamically evolving hidden state variable from noisy observational data.
From condition (i), a circular state-space implies that the underlying filtering task is nonlinear, which precludes the use of popular linear schemes such as the Kalman filter \citep{kalman_new_1960,kalman_new_1961}.
Furthermore, the solution to this so-called circular filtering problem is analytically intractable \citep{kurz_recursive_2016}, and needs to be approximated.
The approximation we derive here goes beyond existing circular filtering algorithms \citep{azmani_recursive_2009,traa_wrapped_2013,kurz_recursive_2014} (see also the review by \citet{kurz_recursive_2016}) by both considering increment observations --- like observations of angular velocity --- and by addressing a continuous stream of observations (condition (ii)).
Furthermore, while these previous approaches changed the representations used between prediction and update steps, ours uses a fixed representation, satisfying our condition (iii).

A recent promising approach to circular filtering \citep{tronarp_continuous-discrete_2018} supports continuous-time state transitions, but is limited to discrete-time and direct (rather than increment) observations.
Their method is based on projection filtering \citep{hanzon_new_1991, brigo_approximate_1999}, a rigorous approach that combines nonlinear filtering with information geometry.
The idea behind projection filtering is to approximate the posterior between consecutive discrete-time observations with a parametric distribution.
This approximation is chosen to minimize the distance between the true and the approximated posterior, as measured by the Fisher metric.
By updating only the values of the parameters, this approach automatically keeps a fixed representation for the posterior in terms of a parametric distribution.
Furthermore, if the approximated posterior and the emission probabilities of the observations are conjugate, updating posterior parameters in light of further discrete-time direct observations is straightforward.
Projection filtering can be generalized to hidden processes that evolve on arbitrary submanifolds of Euclidean space \citep{tronarp_continuous-discrete_2020}, which makes it applicable to the circular filtering problem.
Two challenges hamper the direct application of projection filters to angular path integration.
First, no variants currently exist that handle increment observations, either in discrete or continuous time, as we would require to process angular velocity observations.
In fact, increment observations have generally received little attention in the filtering literature (but see \cite{nusken_state_2019}).
Second, there is currently no framework that combines projection-filtering with angular-valued continuous-time observations.
We will address both challenges in this work.

We introduce a novel continuous-time nonlinear filtering algorithm based on projection filtering, that includes increment observations and can be applied to circular filtering and to angular path integration, and that meets conditions (i)-(iv) outlined above.
To do so, we first describe the general nonlinear filtering problem with observed increment observations in Euclidean space in Section \ref{sec:Setting}.
In Section \ref{sec:Projection filtering for continuous-time increment observations}, we review the projection filtering framework \citep{hanzon_new_1991,brigo_approximate_1999} as an approximate solution for nonlinear filtering, and extend this approach to account for increment observations.
We demonstrate that applying this framework to a \emph{linear} filtering problem recovers the generalized Kalman filter.
In Section \ref{sec: Von Mises Tracking}, we revisit the continuous-time circular filtering problem.
Therein, we first derive a probabilistic algorithm for angular path integration, i.e., when only increment observations are present.
We then account for direct angular-valued observations, in addition to increment observations, by proposing a generative model based on a constant information-rate criterion that supports seamless inclusion into the filtering algorithm.
Combining all of the above, we finally retrieve, as a special case of the general framework, a circular filtering algorithm for Gaussian-type increment and angular direct observations, which we term the circular Kalman filter.
We demonstrate in numerical simulations that this algorithm performs comparably to an asymptotically-exact particle filter, and outperforms a Gaussian approximation in the estimation of both heading direction as well as its associated certainty.


\section{The filtering problem with increment observations}
\label{sec:Setting}

We consider multivariate filtering with observations generated by \emph{increments} of the hidden state, rather than the hidden state itself.
We assume that the hidden state variable $ \mathbf{X}_t \in \mathbb{R}^N $ evolves according to a stochastic differential equation (SDE) of the form:
\begin{align}
	d \mathbf{X}_t & = \mathbf{f}(\mathbf{X}_t,t) \, dt + \Sigma_x^{1/2} \, d\mathbf{W}_t, \label{eq:Setting - dX}
\end{align}
with $ \mathbf{W}_t $ an $ \mathbb{R}^N $-Brownian motion (BM) process, a vector-valued drift function $ \mathbf{f}: \mathbb{R}^N \times  \mathbb{R} \to \mathbb{R}^N $, and a matrix $ \Sigma_x^{1/2} \in  \mathbb{R}^{N \times N} $, which determines the error covariance of the hidden state process.
In the following, we will use the shorthand $ \mathbf{f}_t (\mathbf{x}) := \mathbf{f}(\mathbf{x},t) $ or skip its argument completely (whenever there is no notational ambiguity).
The density $ \tilde{p}_t(\mathbf{x}) := p(\mathbf{X}_t=\mathbf{x}) $ of this stochastic process evolves according to a partial differential equation, the Fokker--Planck equation (FPE):
\begin{align}
	d \tilde{p}_t  = \mathcal{L}^\dagger [\tilde{p}_t] \, dt, \label{eq:Setting - Fokker Planck equation}
\end{align}
with
\begin{align}
\begin{split}
	\mathcal{L}^\dagger [\tilde{p}_t(x)] = & - \sum_{i=1}^N \frac{\partial}{\partial x_i} (f_i(\mathbf{x},t) \tilde{p}_t(\mathbf{x})) \\
	&\quad + \frac{1}{2} \sum_{i,j}^N (\Sigma_x)_{ij} \frac{\partial^2}{\partial x_i \partial x_j}  \tilde{p}_t(\mathbf{x}). \label{eq:Setting - Fokker Planck adjoint}
\end{split}
\end{align}
Equivalently, expectations of a scalar test function $ \phi_t := \phi(\mathbf{X}_t) $ with respect to $ \tilde{p}_t(\mathbf{x}) $ evolve according to
\begin{align}
	 d \mathbb{E}\left[  \phi_t  \right] & =  \mathbb{E} \left[\mathcal{L} [\phi_t] \right]\, dt, \label{eq:Setting - dphi}
\end{align}
with the propagator
\begin{align}
	\mathcal{L} [\phi_t(\mathbf{x},t)] & =  \mathbf{f}_t(\mathbf{x})^T \nabla  \phi_t(\mathbf{x}) + \frac{1}{2} \tr{H_\phi \Sigma_x} , \label{eq:Setting - Fokker Planck operator}	
\end{align}
where $ \nabla $ denotes the gradient with respect to $\mathbf{x}$, $ \tr{\cdot} $ denotes the trace operator and $ H_\phi $ is the Hessian matrix with $ (H_\phi)_{ij} = \frac{\partial^2 \phi_t}{\partial x_i \partial x_j} $.
Note that $ \mathcal{L} $ and $ \mathcal{L}^\dagger $ are adjoint operators with respect to the $ L^2 $ inner product.

We assume that the hidden state process $ \mathbf{X}_t $ in \eqref{eq:Setting - dX} cannot be observed directly, but instead is partially observed through the process $ d\mathbf{U}_t $, which is governed by the infinitesimal state \emph{increments} $ d\mathbf{X}_t $:
\begin{align}
\begin{split}
	d \mathbf{U}_t &= C  \, d\mathbf{X}_t +\Sigma_u^{1/2} d \mathbf{V}_t \\
	& = C \,\mathbf{f}_t(\mathbf{X}_t) \, dt + C \, \Sigma_x^{1/2} d\mathbf{W}_t + \Sigma_u^{1/2} d\mathbf{V}_t. \label{eq:Setting - dY}
\end{split}
\end{align}
where $ \mathbf{V}_t $ is an $ \mathbb{R}^M $-BM process, and $ C \in \mathbb{R}^{M \times N} $.
The matrix $ \Sigma_u \in \mathbb{R}^{M \times M} $ determines the level of noise in the increment observations.
Due to its dependency on the increment $d\mathbf{X}$, the observation process $ d\mathbf{U}_t $ is effectively governed by two noise sources, $d\mathbf{W}_t$ and $d\mathbf{V}_t$.
The first is correlated with the noise in the hidden state dynamics.
The second is independent of it.
This is in contrast to classical filtering problems, which consider the noise in hidden states and observations to be independent (cf.~Appendix \ref{sec:Appendix - standard filtering}).

If $ U_{0:t} = \ \{U_\tau:\tau<t\} $ denotes the filtration generated by the process $ \mathbf{U}_t $, then the filtering problem is to compute the posterior density $ p_t(\mathbf{x}) = p(\mathbf{X}_t = \mathbf{x} | U_{0:t}) $ or, equivalently, the posterior expectations $ \expect{\phi_t} : = \E[\phi(\mathbf{X}_t) | U_{0:t} ] $.
In an uncorrelated noise setting, the Kushner--Stratonovich equation describes the temporal evolution of the posterior expectation $ \expect{\phi_t} $ \cite{stratonovich_conditional_1960,kushner_differential_1964,bain_fundamentals_2009} (Appendix \ref{sec:Appendix - standard filtering}).
To find a similar formal solution for filtering with observed state \emph{increments} (i.e., correlated noise), we can account for the correlations between observations and the hidden state by introducing a slight modification in the Kushner--Stratonovich equation, resulting in a generalized Kushner--Stratonovich equation (gKSE) \citep[Chapter 3.8]{bain_fundamentals_2009} (cf.~\citet{nusken_state_2019}).
For our particular problem, we show in Appendix \ref{sec:Appendix - derivation gKSE} that the dynamics of posterior expectations satisfy
\begin{align}
\begin{split}
d \expect{\phi_t}  = \expect{ \mathcal{L}[\phi_t] } \, dt +&\left(\cov{\phi_t ,\mathbf{f}_t}+ \Sigma_x \expect{ \nabla \phi_t }  \right)^T \\  
&\cdot  C^T \tilde{\Sigma}_u^{-1}  \left( d \mathbf{U}_t - C \expect{\mathbf{f}_t} \, dt \right), \label{eq:Filtering - gKSE}
\end{split}
\end{align}
with $ \cov{\phi_t ,\mathbf{f}_t} = \expect{\phi_t \mathbf{f}_t }  - \expect{\phi_t} \expect{ \mathbf{f}_t }$, and 
\begin{align}
	\tilde{\Sigma}_u = & C \Sigma_x C^T + \Sigma_u.
\end{align}
Note that the right-hand side of \eqref{eq:Filtering - gKSE} does not only depend on $\expect{\phi_t}$, but also on $\expect{\phi_t  \, \mathbf{f}_t}$, $\expect{ \nabla \phi_t }$ and other expectations of potentially nonlinear functions, which in general cannot be computed from $\expect{\phi_t}$ alone.
Therefore, in order to completely characterize the probabilistic solution, we would need one equation \eqref{eq:Filtering - gKSE} for every moment of the posterior, with each moment corresponding to a specific choice of $\phi_t$.
Thus, except for a few very specific generative models, such as linear ones, the dynamics of posterior expectations in \eqref{eq:Filtering - gKSE} will result in a system of an infinite number of coupled SDEs, which in general is analytically intractable.

\begin{remark}
    Equation~\eqref{eq:Filtering - gKSE} is the gKSE if state increments are the only available type of observations.
    In Appendix \ref{sec:Appendix - derivation gKSE}, we extend it to the Kushner--Stratonovich equation when both state increments and (Gaussian-type) direct observations are present.
\end{remark}

In what follows, we will approximately solve the continuous-time filtering problem with observed state increments by projecting the gKSE onto a submanifold of parametric densities with a finite number of parameters, resulting in a \emph{finite} system of coupled SDEs for these parameters.
For this, we will first review the general projection method, which so far has only been applied to classical filtering problems with uncorrelated state and observation noise, and then extend this framework to filtering problems with observed hidden state increments.

\section{Projection filtering for observed continuous-time state increments}
\label{sec:Projection filtering for continuous-time increment observations}

\subsection{The general projection filtering method}

Projection filtering is a method for approximate nonlinear filtering that is based on differential geometry.
In this subsection, we outline briefly the differential geometric setup and derivation of the projection filter, and refer the reader to the seminal papers on projection filters \citep{hanzon_new_1991,brigo_differential_1998,brigo_approximate_1999}, or the intuitive introduction to the subject matter presented in \citep{van_handel_quantum_2005}, for more detailed derivations and in-depth discussion of the method.

In general, we can interpret the solution of a (stochastic) differential equation (such as the FPE \eqref{eq:Setting - Fokker Planck equation}) as a (stochastic) vector field on an infinite-dimensional function space $\mathcal{M}$ of probability density functions $p_t$. 
If the vector field is stochastic (which is usually the case in nonlinear filtering \citep{kushner_differential_1964}), we consider it to be given in Stratonovich form
\begin{align}
	d p_t =& \mathcal{A}^\dagger[p_t]dt+\mathcal{B}^\dagger[p_t]\circ d \mathbf{U}_t, \label{eq:ProFilter - dpt vector field general}
\end{align}
which is the standard choice for stochastic calculus on manifolds.
Let us further assume a parametrization $ p_\theta(\mathbf{x}) := p(\mathbf{x};\theta)$ with a finite set of parameters $ \boldsymbol{\theta} = \{\theta_1, \dots \theta_m \} \in \Theta $, such that the solution to \eqref{eq:ProFilter - dpt vector field general} is reasonably well approximated by these parametrized densities. 

Projection filtering provides a solution to the filtering problem by evolving the parameters $\boldsymbol{\theta}$, thereby constraining the approximate posterior to evolve on a finite-dimensional submanifold $\mathcal{S} = \{ p_\theta(\mathbf{x}); \boldsymbol{\theta} \in \Theta \}$ of $\mathcal{M}$, rather than on $\mathcal{M}$ itself.
This implies that we have to project the vector field in \eqref{eq:ProFilter - dpt vector field general} onto the tangent space $T_{\theta}\mathcal{S}$
\begin{align}
	T_{\theta}\mathcal{S}  = & \text{Span} \left(\frac{\partial p_\theta}{\partial\theta_{1}},\dots,\frac{\partial p_\theta}{\partial\theta_{m}}\right)\subset L^{1}, \label{eq:ProFilter_BasisSubmanifold}
\end{align}
where the $ \frac{\partial p_\theta}{\partial\theta_{i}} $ denote the basis vectors of this tangent space.
Intuitively, an orthogonal projection minimizes at each timestep the distance between the true posterior $ p_t $ and its approximation $ p_\theta $ with respect to a Riemannian metric which, for probability distributions, corresponds to the Fisher metric \citep{amari_differential-geometrical_1990}
\begin{align}
	g_{ij} &=  \E_{p_\theta} \left[ \frac{\partial \log p_\theta(\mathbf{x})}{\partial \theta_i}  \frac{\partial \log p_\theta(\mathbf{x})}{\partial \theta_j} \right]. \label{eq:ProFilter - Fisher metric}
\end{align}
This allows us to use the general orthogonal projection formula
\begin{align}
	\Pi_{\theta} [\mathcal{Z}] & = \sum_{i}\sum_{j}g^{ij} \innerproduct{\mathcal{Z},\frac{\partial p_\theta}{\partial\theta_{i}}}_\theta \cdot \frac{\partial p_\theta}{\partial\theta_{j}},\label{eq:ProFilter - OrthogonalProjection}
\end{align}
where $\Pi_{\theta}$ denotes the projection operator, the $g^{ij} $ are the components of the inverse Fisher metric, and
\begin{align}
	\innerproduct{Z_1,Z_2}_\theta := & \int_\Omega d\mathbf{x}\,\frac{Z_1(\mathbf{x}) Z_2 (\mathbf{x})}{p_\theta(\mathbf{x})}, \ \ \ Z_1,Z_2\in T_{\theta}\mathcal{S} .  \label{eq:ProFilter - scalar product}
\end{align}
is the inner product that is associated with the Fisher metric (see Proposition \ref{proposition: scalar product and Fisher metric} in Appendix \ref{sec:Apx - Fisher metric}).

To find the parameter updates resulting from this projection, we apply the projection \eqref{eq:ProFilter - OrthogonalProjection} to the dynamics of the probability density $ p_t $ in \eqref{eq:ProFilter - dpt vector field general}.
Since our approximate dynamics evolve along tangent vectors to the manifold parametrized by $ \boldsymbol{\theta} $, the posterior $ p_\theta $ will stay on this manifold.
Hence, the left-hand side of \eqref{eq:ProFilter - dpt vector field general} can be written in terms of the basis vectors of $T_{\theta}\mathcal{S}$ by using the chain rule:
\begin{align}
	\Pi_{\theta}[dp_\theta]=\sum_{j}\frac{\partial p_\theta}{\partial\theta_{j}}d\theta_{j}.
\end{align}
Further, by letting the projection act on the right-hand side of this equation, and consecutively comparing coefficients in front of the basis vectors $\frac{\partial p_\theta}{\partial\theta_{j}}$, we find the following Stratonovich SDEs for the parameters of the projected density following the evolution in \eqref{eq:ProFilter - dpt vector field general}:
\begin{align}
\begin{split}
	d\theta_{j} = &  \sum_{i} g^{ij} \innerproduct{ \mathcal{A}^\dagger [p_\theta],\frac{\partial p_\theta}{\partial\theta_{i}}}_\theta \,dt  \\
	& +\sum_{i} g^{ij} \innerproduct{\mathcal{B}^\dagger [p_\theta],\frac{\partial p_\theta}{\partial\theta_{i}}}_\theta \circ d \mathbf{U}_t. \label{eq:ProFilter - StratSDE_Parameters daggar} 
\end{split}
\end{align}
This is the result of \citet[Theorem 4.3]{brigo_approximate_1999}.

To facilitate the comparison to the gKSE \eqref{eq:Filtering - gKSE}, we further slightly rewrite this SDE:
\begin{align}
\begin{split}
	d \theta_j& = \sum_{i} g^{ij} \bigg( \int_\Omega dx \, \mathcal{A}^\dagger[p_\theta] \frac{\partial \log p_\theta}{\partial \theta}\bigg) \,dt  \\
	& \quad + \sum_{i} g^{ij}  \bigg( \int_\Omega dx\,  \mathcal{B}^\dagger[p_\theta] \frac{\partial \log p_\theta}{\partial \theta}\bigg)\circ d \mathbf{U}_t 
\end{split} \\
\begin{split}
	&  =  \sum_{i} g^{ij} \expecttwo{ \mathcal{A} \left[\frac{\partial \log p_\theta}{\partial\theta_{i}}\right] }{\theta} \, dt  \\
	& \quad + \sum_{i} g^{ij}  \expecttwo{\mathcal{B} \left[\frac{\partial \log p_\theta}{\partial\theta_{i}}\right]}{\theta}\circ d \mathbf{U}_t  ,\label{eq:ProFilter - StratSDE_Parameters} 
\end{split}
\end{align}
where $ \expecttwo{\cdot}{\theta} $ here denotes the expectation with respect to the projected density $ p_\theta $, and $ \mathcal{A} $ and $ \mathcal{B} $ denote the adjoint of $ \mathcal{A}^\dagger $ and $ \mathcal{B}^\dagger $ with respect to the $ L^2 $ scalar product.
Rewriting the SDE in such a way allows us to immediately identify the operators $\mathcal{A}$ and $ \mathcal{B} $ on the right-hand side of this equation with the operators used for propagating expectations rather than densities.

To illustrate how to identify the operator $\mathcal{A}$ concretely, let us consider the filtering problem \emph{without} any observations, i.e.,~a simple diffusion, which is formally solved by the Fokker--Planck Equation \eqref{eq:Setting - Fokker Planck equation}.
Noting that the operator $ \mathcal{L}^\dagger $ propagates the density through time, we identify  $ \mathcal{A}^\dagger = \mathcal{L}^\dagger $.
Similarly, we find $ \mathcal{A} = \mathcal{L} $ for the adjoint.
Thus, the projection can be immediately determined from the time evolution of the expectation in Eq.~\eqref{eq:Setting - dphi}, with $ \phi_t = \frac{\partial \log p_\theta}{\partial \theta_{i}} $:
\begin{align}
	d \theta_j &= \sum_{i} g^{ij} \expect{\mathcal{L}\left[\frac{\partial \log p_\theta}{\partial \theta_{i}}\right]} \, dt. \label{eq:ProFilter - diffusion_Example} 
\end{align}

If state increments are observed, expectations are propagated by using the gKSE \eqref{eq:Filtering - gKSE}, and identification of the operators $\mathcal{A}$ and $\mathcal{B}$ is possible after having transformed the gKSE to Stratonovich form, as we will see in the next section.
Since this is usually easier to do than transforming the generalized Kushner equation for the posterior \emph{density} in Stratonovich form, Eq.~\eqref{eq:ProFilter - StratSDE_Parameters} is a more convenient choice for the parameter dynamics than Eq.~\eqref{eq:ProFilter - StratSDE_Parameters daggar}.

\begin{remark}
	The derivation in the seminal papers on projection filtering \citep{hanzon_new_1991,brigo_approximate_1999} follows a slightly different route but leads to the same result \eqref{eq:ProFilter - StratSDE_Parameters daggar}.
	We also refer the reader to the very accessible derivation presented in \citep{van_handel_quantum_2005}.
\end{remark}

\subsection{Projection with observed state increments}

As we have seen, the adjunction between the SDE evolution operators for densities and associated expectations allowed us to use the expectation's evolution equation to derive the projection filter for a specific problem \eqref{eq:ProFilter - diffusion_Example}. 
The same applies if the evolution of the density (or equivalently, that of the expectations) is a \emph{stochastic} differential equation, as is the case for the KSE and the gKSE \eqref{eq:Filtering - gKSE}, as long as these are given in Stratonovich form.
This allows us to formulate the projection filter for filtering with observed state increments:

\begin{theorem}
	\label{theorem: Projection filter}
	The projection filter for the filtering problem with observed state increments is given by the following SDE of the parameters $ \boldsymbol{\theta} $ of a projected density $ p_\theta $:
	
	\small
	\begin{align}
	\begin{split}
		d \theta_{j}  = & \sum_{i} g^{ij} \bigg[ \expecttwo{\tilde{\mathcal{L}}\left[ \frac{\partial \log p_\theta}{\partial \theta_{i}} \right]}{\theta} \\
		& \quad - \frac{1}{2} \bigg( \expecttwo{ || \tilde{\Sigma}_u^{-1/2} C \, \mathbf{f}_t||^2 \frac{\partial \log p_\theta}{\partial \theta_{i}} }{\theta}  \\
		& \quad +  \expecttwo{\tr{\tilde{\Sigma}_u^{-1} C  J_f \Sigma_x C^T} \frac{\partial \log p_\theta}{\partial \theta_{i}} }{\theta} \bigg)\bigg] \, dt  \\
		& + \sum_{i} g^{ij}  \bigg( \expecttwo{\frac{\partial \log p_\theta}{\partial \theta_{i}}  \mathbf{f}_t }{\theta}  + \Sigma_x\expecttwo{ \nabla \frac{\partial \log p_\theta}{\partial \theta_{i}}  }{\theta} \bigg)^T  \\
		& \quad  \quad \cdot C^T \tilde{\Sigma}_u^{-1}   \circ d \mathbf{U}_t, \label{eq:ProFilter - increment observations}
	\end{split}
	\end{align}
	\normalsize
	with $ (J_f)_{ij} = \frac{\partial f_i}{\partial x_j} $ denoting the Jacobian matrix, and with the short-hand modified generator
	\begin{align}
		\begin{split}
		\tilde{\mathcal{L} } [\phi_t] := &  \big(  \nabla  \phi_t(\mathbf{x}) \big)^TC^{-1} \Sigma_u \tilde{\Sigma}_u^{-1} C \,  \mathbf{f}_t(\mathbf{x})  \\
		&  + \frac{1}{2} \text{Tr}( H_\phi \Sigma_x C^{-1} \Sigma_u \tilde{\Sigma}_u^{-1} C).
		\end{split}
	\end{align}
\end{theorem}

\begin{proof}
	As a first step, let us rewrite the gKSE \eqref{eq:Filtering - gKSE} in Stratonovich form (Corollary \ref{corollary: gKSE Stratonovich} in \ref{sec:Appendix - derivation gKSE}): 
	\begin{align}
	\begin{split}
		d \expect{\phi_t}   = &  \bigg[ \expect{\tilde{\mathcal{L}}[\phi_t]}  - \frac{1}{2} \bigg(  \cov{\phi_t, ||\tilde{\mathbf{\Sigma}}_u^{-1/2} C \, \mathbf{f}_t ||^2 }  \\
		& \quad + \tr{\tilde{\Sigma}_u^{-1} C  \,  \cov{\phi_t,J_f} \Sigma_x C^T} \bigg) \bigg] \, dt  \\
		&  + \big[ \cov{\phi_t, \mathbf{f}_t}  + \Sigma_x\expect{ \nabla \phi_t} \big]^T C^T \tilde{\Sigma}_u^{-1}  \circ d\mathbf{U}_t . \label{eq:ProFilter - gKSE in Stratonovich}
	\end{split}
	\end{align}
	This equation allows us to identify the operators $ \mathcal{A} $ and $ \mathcal{B} $ in Eq.~\eqref{eq:ProFilter - StratSDE_Parameters} as the operations acting on $ \phi_t $  in front of $ dt $ and $ d \mathbf{U}_t $, respectively.
	By letting these operators act on $ \phi_t = \frac{\partial \log p_\theta}{\partial \theta_j} $ in Eq.~\eqref{eq:ProFilter - StratSDE_Parameters}, and evaluating the expectations under the projected density $ p_\theta $, we obtain SDEs for the desired parameters.
	These can be further simplified by using
	\begin{align}
		\expecttwo{\frac{\partial \log p_\theta}{\partial \theta_j}}{\theta} & = \frac{\partial}{\partial \theta_j} \int_\Omega dx\,  p_\theta(x) = 0, \nonumber
	\end{align}
	which yields \eqref{eq:ProFilter - increment observations}.
\end{proof}

The 1D special case follows directly from \eqref{eq:ProFilter - increment observations}.
\begin{corollary}
	For univariate filtering problems, i.e.~a filtering problem with $ N=M=1 $ in \eqref{eq:Setting - dX} and \eqref{eq:Setting - dY}, with $ C = c $, $ \Sigma_x = \sigma_x^2 $, $ \Sigma_u = \sigma_u^2 $ and $ \tilde{\Sigma}_u = \tilde{\sigma}_u^2 = c^2 \sigma_x^2 + \sigma_u^2 $, the projection filter with observed state increments reads:
	\small
	\begin{align}
	\begin{split}
		d \theta_{j}  = & \sum_{i} g^{ij} \bigg[\frac{\sigma_u^2}{\tilde{\sigma}_u^2 }\expecttwo{\mathcal{L}\left[ \frac{\partial \log p_\theta}{\partial \theta_{i}} \right]}{\theta}  \\
		& \quad  - \frac{c^2}{2 \tilde{\sigma}_u^2} \bigg( \expecttwo{f_t^2 \frac{\partial \log p_\theta}{\partial \theta_{i}} }{\theta} + \sigma_x^2 \expecttwo{\frac{\partial f_t}{\partial x} \cdot \frac{\partial \log p_\theta}{\partial \theta_{i}}}{\theta} \bigg)\bigg] \, dt  \\
		+ &  \sum_{i} g^{ij}  \bigg[ \frac{c}{\tilde{\sigma}_u^2} \bigg( \expecttwo{ f_t  \frac{\partial \log p_\theta}{\partial \theta_{i}} }{\theta}  \\
		& \quad + \sigma_x^2  \expecttwo{\frac{\partial}{\partial x}   \frac{\partial \log p_\theta}{\partial \theta_{i}} }{\theta}   \bigg) \bigg] \circ d U_t. \label{eq:ProFilter - increment observations 1D}
	\end{split}
	\end{align}
	\normalsize
\end{corollary}

\subsection{Projection on exponential family distributions}

Analogous to \citep{brigo_approximate_1999}, it is possible to derive explicit filter equations for the natural parameters of a projected exponential family distribution.
Consider the following exponential family parametrization:
\begin{align}
	p_\theta(\mathbf{x}) := \exp (\boldsymbol{\theta}^\top \mathbf{T}(\mathbf{x}) - \Psi(\boldsymbol{\theta})), \label{eq:ProFilter - exponential family natural}
\end{align}
where $ \boldsymbol{\theta} $ is the vector of \emph{natural} or \emph{canonical parameters}, $ \mathbf{T}(x) $ is the vector of sufficient statistics and $ \exp (\Psi(\boldsymbol{\theta})) $ is the normalization.
\begin{corollary}
	\label{corollary: Projection filter for exponential families}
	 The projection filter for the filtering problem with observed state increments is given by the following SDE of the natural parameters $ \boldsymbol{\theta} $ of a projected density $ p_\theta $ belonging to the exponential family:
	\begin{align}
	\begin{split}
		d \theta_{j}  = & \sum_{i} g^{ij} \bigg[ \expecttwo{\tilde{\mathcal{L}}\left[T_i(\mathbf{x}) \right]}{\theta}  \\
		& \quad - \frac{1}{2} \bigg( \expecttwo{ || \tilde{\Sigma}_u^{-1/2} C \, \mathbf{f}_t||^2 (T_i(\mathbf{x}) - \eta_i) }{\theta}  \\
		& \quad +  \expecttwo{\tr{\tilde{\Sigma}_u^{-1} C  J_f \Sigma_x C^T}(T_i(\mathbf{x}) - \eta_i)}{\theta} \bigg)\bigg] \, dt  \\
		 +&  \sum_{i} g^{ij}  \big( \expecttwo{ (T_i (\mathbf{x})- \eta_i) \mathbf{f}_t }{\theta}  + \Sigma_x \expecttwo{ \nabla T_i (\mathbf{x})}{\theta} \big)^T  \\
		& \quad  \quad \cdot C^T \tilde{\Sigma}_u^{-1}   \circ d \mathbf{U}_t, \label{eq:ProFilter - Natural Parameters} 
	\end{split}
	\end{align}
	where $ \eta_i = \expecttwo{T_i(\mathbf{x})}{\theta} $ is the posterior expectation of the sufficient statistic $ T_i(\mathbf{x}) $.
	The parameters $ \boldsymbol{\eta} $ are sometimes referred to as dual or expectation parameters.
\end{corollary}
\begin{proof}
	Making use of the duality relation between natural and expectation parameters for exponential families,
	\begin{align}
		\frac{\partial}{\partial \theta_i} \log p_\theta (\mathbf{x}) & = T_i(\mathbf{x}) - \frac{\partial}{\partial \theta_i} \Psi(\boldsymbol{\theta}) = T_i(\mathbf{x}) -\eta_i,
	\end{align}
	and the fact that $ \frac{\partial}{\partial x} \eta_i = 0 $, Eq.~\eqref{eq:ProFilter - Natural Parameters} follows directly from the projection filter with observed state increments \eqref{eq:ProFilter - increment observations}.
\end{proof}

\begin{example}[Generalized Kalman--Bucy filter]
	In order to demonstrate the general approach, let us consider a model with linear state dynamics
	\begin{align}
		dX_t & = a X_t dt + \sigma_x \, dW_t, \label{eq:gKBF - dX} \\
		d U_t & = c\,  dX_t + \sigma_u \, dV_t. \label{eq:gKBF - dY}
	\end{align}
	Here, we will show that a projection on a Gaussian manifold with 
	\begin{align}
		p_\theta(x) & = \mathcal{N} (x; \mu_t, \sigma_t^2 )
	\end{align}
	results in dynamics for the parameters $ \mu_t $ and $ \sigma_t $ that are consistent with the generalized Kalman--Bucy Filter for observed state increments \citep[Section 4.2]{nusken_state_2019}.
	In fact, since Eqs.~\eqref{eq:gKBF - dX} and \eqref{eq:gKBF - dY} are both linear, the posterior density is a Gaussian and thus the projection filter becomes exact.
	For this particular problem, the projection filter reads:
	
	\small
	\begin{align}
	\begin{split}
		d \theta_{j}  = &  \sum_{i} g^{ij} \bigg[\frac{\sigma_u^2}{\tilde{\sigma}_u^2 }\expecttwo{ a X_t \frac{\partial}{\partial x} \frac{\partial \log p_\theta}{\partial \theta_i} + \sigma_x^2  \frac{\partial^2}{\partial x^2} \frac{\partial \log p_\theta}{\partial \theta_i} }{\theta}  \\
		& \quad  - \frac{c^2 a^2}{2 \tilde{\sigma}_u^2}  \expecttwo{X_t^2 \frac{\partial \log p_\theta}{\partial \theta_{i}} }{\theta} \bigg] \, dt  \\
		 + & \sum_{i} g^{ij}  \bigg[ \frac{c}{\tilde{\sigma}_u^2} \bigg( a \expecttwo{ X_t \frac{\partial \log p_\theta}{\partial \theta_{i}} }{\theta}  \\
		& \quad +\sigma_x^2  \expecttwo{\frac{\partial}{\partial x}   \frac{\partial \log p_\theta}{\partial \theta_{i}} }{\theta}    \bigg) \bigg] \circ d U_t. \label{eq:ProFilter_1Dlinear}
	\end{split}
	\end{align}
	\normalsize
	We will use this to determine SDEs for the parameters $ \mu_t $ and $ \sigma_t^2 $, with $ \frac{\partial \log p_\theta}{\partial \mu_t} = \frac{x-\mu_t}{\sigma_t^2} $ and $ \frac{\partial \log p_\theta}{\partial (\sigma_t^2)} = -\frac{1}{2 \sigma_t^2} + \frac{(x-\mu_t)}{2 \sigma_t^4} $, respectively, under the Gaussian assumption.
	First, the components of the Fisher information matrix of a Gaussian parametrized by its expectation parameters are given by
	\begin{align}
		g_{\mu \mu}  = \frac{1}{\sigma_t^2}, \,\,\,\  g_{\sigma^2 \sigma^2}  = \frac{1}{2 \sigma_t^4}, \,\,\,\ 	g_{\sigma^2 \mu}  = g_{\mu \sigma^2} = 0.
	\end{align}
	Since the matrix is diagonal, the components of its inverse are $ g^{\mu \mu} = g_{\mu \mu}^{-1} = \sigma_t^2$ and $ g^{\sigma^2 \sigma^2 }= g_{\sigma^2 \sigma^2 }^{-1} = 2 \sigma_t^2$.
	This considerably simplifies the projection in Eq.~\eqref{eq:ProFilter_1Dlinear} for the time evolution of $ \mu_t $ and $ \sigma_t $.
	Explicitly carrying out the expectations in Eq.~\eqref{eq:ProFilter_1Dlinear} under the assumed Gaussian density, the SDEs for these parameters read:
	\begin{align}
		d \mu_t & = a \mu_t \,dt + \frac{c}{c \sigma_x^2 + \sigma_u^2}(a \sigma_t^2 + \sigma_x^2) \cdot  (d U_t - a c \mu_t \, dt), \label{eq:gKBF - dmu} \\
		d \sigma_t^2 & = \left[ 2 a \sigma_t^2 + \sigma_x^2 - \frac{c^2  }{c \sigma_x^2 + \sigma_u^2} ( a \sigma_t^2 + \sigma_x^2 )^2 \right] \, dt. \label{eq:gKBF - dsig2}
	\end{align}
	We found these It\^{o} SDEs from their Strontonovich form by noting for the first line that the quadratic variation between $ \sigma_t $ and the observations process $ U_t $ is zero.
	In other words, no correction term according to the Wong--Zakai theorem \citep{wong_relation_1965} is required, such that both It\^{o} and Stratonovich form have the same representation.
	Note that for a nonlinear generative model, this will in general \emph{not} be the case \cite{brigo_approximate_1999}.
	Equations \eqref{eq:gKBF - dmu} and \eqref{eq:gKBF - dsig2} are identical to the generalized Kalman--Bucy filter \cite[Eqs. (62) \& (65)]{nusken_state_2019}, thus demonstrating the validity of our approach.
\end{example}

Despite being able to reproduce existing results, the main purpose of the projection filtering approach is to simplify potentially hard filtering problems such that the parameter SDEs become analytically accessible.
This becomes particularly useful if a certain parametric form of the posterior is desired, for instance for computational reasons, and can be very appealing if expectations are easily carried out under the assumed posterior and the Fisher matrix is straightforward to invert or even diagonal.

\begin{example}[Multivariate Gaussian with diagonal covariance matrix]
	From a computational perspective, projection on a Gaussian density with diagonal covariance matrix can be advantageous in certain situations.
	In particular, such a solution only requires equations for $ 2N $ parameters, instead of $ N^2 +N $ for a general Gaussian with Fisher matrix components
	\begin{align}
		& g_{\mu_i,\mu_j} = (\Sigma_t^{-1})_{ij}, \quad g_{\mu_i,\sigma^2_{nm}} = 0, \nonumber \\
		& g_{ \sigma_{ij}, \sigma{nm} } = \frac{1}{4} \left[ (\Sigma_t^{-1})_{in} (\Sigma_t^{-1})_{jm} + (\Sigma_t^{-1})_{im} (\Sigma_t^{-1})_{jn}  \right], \nonumber
	\end{align}
	which is in general hard to invert.
	For a Gaussian with diagonal covariance matrix, these simplify to 
	\begin{align}
		g_{\mu_i,\mu_i} = (\Sigma_t^{-1})_{ij} , \quad g_{ \sigma^2_{ii}, \sigma^2{ii} }  = \frac{1}{2} (\Sigma_t^{-1})_{ii}^2  = \frac{1}{2 \sigma_{ii}^2} , \nonumber
	\end{align}
	while all other components evaluate to zero, making the Fisher matrix diagonal and straightforward to invert.
	Since the diagonality of the covariance matrix effectively decouples the dimensions, expectations can be carried out in each dimension separately.
	Nevertheless, the specific form of the parameter SDEs will crucially depend on the specific form of the nonlinear function $ \mathbf{f}_t(\mathbf{x}) $.
	For instance, considering the linear case, i.e.~$ \mathbf{f}_t(\mathbf{x}) = A \mathbf{x} $, yields
	
	\small
	\begin{align}
		d \boldsymbol{\mu}_t & = A\boldsymbol{\mu}_t + \left( \Sigma_x + \mathrm{diag}(\sigma_{ii}^2) A \right) C^T \tilde{\Sigma}_u^{-1}  ( d \mathbf{U}_t - C A \boldsymbol{\mu}_t \, dt ), \nonumber \\
		\frac{d \sigma_{ii}^2}{dt} & = 2 \sigma_{ii}^2 (A - \Sigma_x C^T \tilde{\Sigma}_u^{-1} C A)_{ii}  \nonumber \\
		& - 2 \sigma_{ii}^4 ( A^T C^T \tilde{\Sigma}_u^{-1} C A )_{ii} + (\Sigma_x - \Sigma_x C^T \tilde{\Sigma}_u^{-1} C \Sigma_x)_{ii}. \nonumber
	\end{align}
	\normalsize
\end{example}


\section{Continuous-time circular filtering}
\label{sec: Von Mises Tracking}

In this section, we will consider continuous-time circular filtering with observed state increments as a concrete application of the framework derived above.
We will further extend it to account for quasi continuous-time von Mises-valued observations (formally defined below), to provide a continuous-time generalization of the discrete-time circular filtering problem that is frequently encountered in spatial navigation problems.

\subsection{Assuming observed angular increments only}

We assume that the hidden state $ X_t $ is parametrized on $ S^1 $ by angle $ \varphi_t \in [0, 2\pi)  $, effectively embedding $ S^1 $ in $ \mathbb{R}^2 $ as a unit circle.
We further assume that $\varphi_t$ follows a diffusion on the circle:
\begin{align}
	d \varphi_t & =  f(\varphi_t,t) \, dt + \sigma_\varphi \, dW_t,  \label{eq:d varphi}
\end{align}
where $ W_t $ is now an $ \mathbb{R}^1 $-BM process, and drift and diffusion functions are as defined in Section \ref{sec:Setting}.
The propagator $ \mathcal{L}[\cdot] $ for this process is the same as for the corresponding process in $ \mathbb{R}^1 $ as given in Eq.~\eqref{eq:Setting - Fokker Planck operator}.

For state processes that evolve on submanifolds of $ \mathbb{R}^n $, as the $S^1$ considered here, \citet{tronarp_continuous-discrete_2020} have shown that projection filter equations are identical to the case where the state variable $\mathbf{X}_t$ evolves in Euclidean space.
Since the mathematical operations performed to derive the projection filter with observed state increments in Theorem \ref{theorem: Projection filter} are essentially the same as in \citet{tronarp_continuous-discrete_2020} for the state diffusion, their result carries over to our problem.
Thus, Theorem \ref{theorem: Projection filter} can straightforwardly be applied to the circular filtering problem by considering a circular projected density $ p_{\theta}(\varphi) $, such as the von Mises or a wrapped normal distribution.

\begin{example}[Circular diffusion with observed state increments]
	\label{sec:Example - circular diffusion increment obs}
	In this example, we explicitly model angular path integration as the estimation of a circular diffusion based on observed angular increments.
	Consider a model where the hidden state evolves according to a Brownian motion on the circle, with noisy observations of its increment
	\begin{align}
		d \varphi_t = & \frac{1}{\sqrt{\kappa_\varphi}} \, dW_t, \label{eq:CircFilt - dphi example} \\
		d U_t = & d \varphi_t + \frac{1}{\sqrt{\kappa_u}} \, dV_t. \label{eq:CircFilt - dY example}
	\end{align}
	Here, $ \varphi_t $ could, for instance, correspond to the heading direction of an animal (or a robot) that is navigating in darkness and only has access to self-motion cues $ d U_t $, i.e.,~measurements of angular increments, but not to direct heading cues such as landmark positions.
	We chose to parametrize the diffusion constants in terms of precisions, $\kappa_\varphi$ and $\kappa_u$, to make units comparable to that of the precision of the projected density, which we will denote $\kappa_t$.
	Thus, the parameter $\kappa_\varphi$ governs the speed of the hidden state diffusion, and the parameter $\kappa_u$ modulates the reliability of the observation process that is governed by the increments.
	The gKSE for this model's posterior expectation of a test function $ \phi_t:=\phi(\varphi_t) $ in Stratonovich form reads:
	
	\small
	\begin{align}
		d \expect{\phi_t} & = \frac{1}{2 (\kappa_\varphi+\kappa_u)} \expect{\frac{\partial^2}{\partial \varphi^2} \phi_t} \, dt + \frac{\kappa_u}{\kappa_\varphi+\kappa_u} \expect{\frac{\partial}{\partial \varphi} \phi_t } \circ d U_t. \label{eq:CircFilt - KSE circular simple}
	\end{align}
	\normalsize
	As a result, the projection filter for the parameters $ \boldsymbol{\theta} $ becomes (cf.~Eq.~\eqref{eq:ProFilter - increment observations})
	\begin{align}
	\begin{split}
		d \theta_j  = \frac{1}{\kappa_\varphi + \kappa_u} \sum_{i} g^{ij}  & \Big[ \frac{1}{2} \expecttwo{ \frac{\partial^2 }{\partial \varphi^2} \frac{\partial \log p_\theta}{\partial \theta_i} }{\theta} \, dt  \\ 
		& + \kappa_u \expecttwo{ \frac{\partial}{\partial \varphi} \frac{\partial \log p_\theta }{\partial \theta_i} }{\theta} \circ d U_t \Big]. \label{eq:CircFilt - dtheta simple}
	\end{split}
	\end{align}
	
	We now want to solve the circular filtering problem with observed state increments by projecting on the von Mises density
	\begin{align}
		p_{\mu,\kappa} (\varphi) = \mathcal{VM}(\varphi; \mu_t, \kappa_t) = \frac{1}{2 \pi I_0(\kappa_t)} \exp \left( \kappa_t \cos (\varphi - \mu_t) \right), \label{eq:CircFilt - vonMises density}
	\end{align} 
	parametrized by mean $ \mu_t $ and precision $ \kappa_t $, using Eq.~\eqref{eq:CircFilt - dtheta simple}.
	Unlike e.g.,~the wrapped normal distribution, which is another popular choice for unimodal circular distributions, the von Mises distribution is an exponential family distribution and could alternatively be written in natural parametrization \eqref{eq:ProFilter - exponential family natural}. Here, we chose to parametrize it by $ \mu_t $ and $ \kappa_t $, as it significantly simplifies the computation of the Fisher metric and its inverse, which appears on the right-hand side of \eqref{eq:CircFilt - dtheta simple}.
	Noting that
	\begin{align}
		\frac{\partial }{\partial \mu_t} \log p_{\mu,\kappa} (\varphi) = &  \kappa_t \sin ( \varphi - \mu_t), \label{eq:vonmises dlogp d mu}\\
		\frac{\partial }{\partial \kappa_t} \log p_{\mu,\kappa} (\varphi) = & - F(\kappa_t) + \cos(\varphi - \mu_t), \label{eq:vonmises dlogp d kappa}
	\end{align}
	where $F(\kappa) = \frac{I_1(\kappa)}{I_0(\kappa)}$
	denotes a ratio of Bessel functions, the components of the Fisher metric (with respect to the $ \mu_t $, $ \kappa_t $ parametrization) are given by:
	\begin{align}
		g_{\mu \mu}   &=  \kappa_t^2 \, \expecttwo{ \sin^2 (\varphi_t-\mu_t)}{\mu,\kappa}= \kappa_t \, F(\kappa_t), \\
		\begin{split}
		g_{\kappa \kappa}  & =  \expecttwo{ \left( F(\kappa_t) + \cos(\varphi_t - \mu_t) \right)^2}{\mu,\kappa} \\
		& =  1- \frac{F(\kappa_t)}{\kappa_t}  - F(\kappa_t)^2, 
		\end{split}\\
		g_{\mu \kappa} & = g_{\kappa \mu} = 0.
	\end{align}
	Since the Fisher metric is diagonal, the components of its inverse are simply $ g^{\mu \mu} = g_{\mu \mu}^{-1} $, $ g^{\kappa \kappa} = g_{\kappa \kappa}^{-1} $ and $ g^{\mu \kappa} = g^{\kappa \mu}=0 $.
	
	Using these $ g_{ij} $'s and explicitly computing the expectations on the right-hand side of Eq.~\eqref{eq:CircFilt - dtheta simple} with respect to the von Mises approximation, we find the projection filter equations for this model:
	\begin{align}
		d \mu_t = &  \frac{\kappa_u}{\kappa_\varphi+\kappa_u}  \cdot d \mathbf{U}_t, \label{eq:CircFilt - dmu}\\
		d \kappa_t = & - \frac{1}{2(\kappa_\varphi+\kappa_u)} \mathcal{F}(\kappa_t) \, dt ,\label{eq:CircFilt - dkappa}
	\end{align}
	where we defined the strictly positive function
	\begin{align}
		\mathcal{F}(\kappa_t) & = \frac{F(\kappa_t)}{1-\frac{F(\kappa_t)}{\kappa_t} - F(\kappa_t)^2 },
	\end{align}
	and found the It\^{o} form of $d \mu_t$ from its Stratonovich form by noting that the noise variance is constant, making this conversion straightforward.
	
	The projection filter defined by Eqs.~\eqref{eq:CircFilt - dmu} and \eqref{eq:CircFilt - dkappa} for orientation tracking in darkness has an intuitive interpretation:
	the mean $ \mu_t $ is updated according to the angular increment observations, weighted by their reliability, as quanitified by $ \kappa_u $.
	Even in the presence of such observations, the estimate's precision, $ \kappa_t $, decays towards zero, since $ \mathcal{F}(\kappa_t) $ is strictly positive.
	This decay reflects the accumulation of noisy observations.
	Very informative angular velocity observations with large $ \kappa_u $ may slow the decay, but cannot fully prevent it.
	In other words, without direct angular observations (which we will introduce in Section \ref{sec:Quasi continuous-time von Mises observations} below), the estimate will inevitably become less accurate over time.
	
\begin{figure*}
	\centering
	\includegraphics[width=0.9\linewidth]{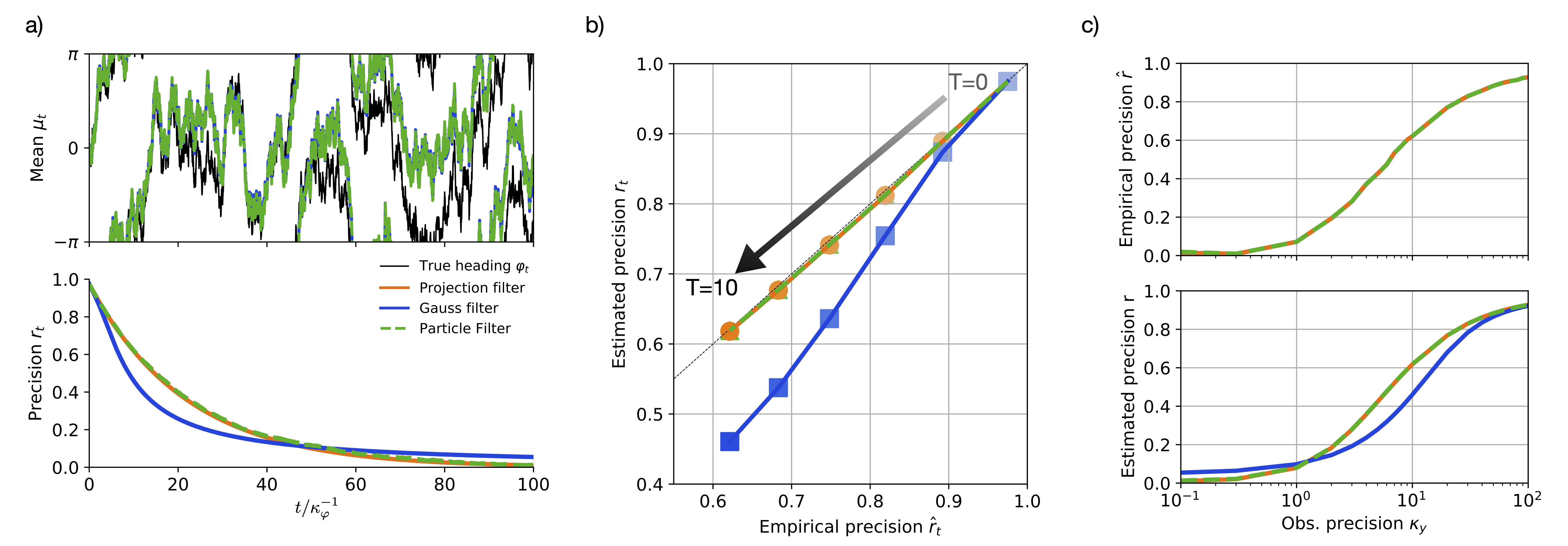}
	\caption{ 
		{\bf{Circular filtering with observed angular increments.} }
		a) The von Mises projection filter in Eqs.~\eqref{eq:CircFilt - dmu} and \eqref{eq:CircFilt - dkappa} is able to track a diffusion process on the circle based on observed increments, with mean $ \mu_t $ and precision $ r_t $ (given by $ r_t = \frac{I_1(\kappa_t)}{I_0(\kappa_t)} $) matching that of a particle filter (curves are on top of each other).
		b) While the precision $ \hat{r}_t $ estimated from the deviation between mean $ \mu_t $ and true trajectory $ \varphi_t $ and the precision $ r_t $ estimated by the filter coincide for both the von Mises projection filter and the particle filter, the Gaussian projection filter (``Gauss filter'') tends to underestimate its precision. 
		c) Empirical (upper panel) and estimated (lower panel) precision for different values of the observations precision $ \kappa_u $ at time $ T=10\  \kappa_\varphi^{-1} $. 
		Note that in the upper panel, the empirical precision $ \hat{r}_t $ of the different filters is identical.
		Parameters for a) and b) are $ \kappa_\varphi = 1 $, $ \kappa_u = 10 $, times are in units of $ \kappa_\varphi^{-1} $.
		Simulations in b) and c) were averaged over 5000 runs.
	}
	\label{fig:Increment-Observations}
\end{figure*}

	In Figure \ref{fig:Increment-Observations}a, we illustrate in an example simulation that, despite the presence of angular increment observations, the estimate slowly drifts away from the true heading $ \varphi_t $.
	As a benchmark we use a particle filter, and further compare mean $\mu_t $ and precision $ r_t = F(\kappa_t) $ of the projection filter to that estimated by a Gaussian projection filter approximation (see Appendix \ref{sec:Apx - numerics} for details on these benchmarks).
	Such a filter relies on the assumption that the hidden state $\varphi_t$ evolves on the real line, and thus leads to a slight deviation in the dynamics of the estimated precision $\kappa_t$.
	
	Numerically, our projection filter's performance in this example is indistinguishable from that of the particle filter (Figure \ref{fig:Increment-Observations}b, c), and its estimated precision $ r_t $ matches exactly the empirical precision evaluated by averaging the estimation error over 5000 simulation runs.
	The estimated precision of the Gaussian approximation, in contrast, systematically underestimates its precision for large observation reliability, and overestimates it when the angular increment observations become very noisy (Figure \ref{fig:Increment-Observations}c).
\end{example}

\begin{example}[Higher-order circular distributions]
\label{sec:example Higher-order circular distributions}
	The previous example is one of the simplest examples of a circular filtering problem, and results in an approximated posterior that is always unimodal.
	This might be insufficient for certain settings in which we would like to consider more sophisticated projected densities.
	To show that our framework extends beyond such simple models, let us consider a class of circular distributions with exponential-family densities of the form
	\begin{align}
		p(\varphi) & = \frac{1}{Z(\mathbf{a},\mathbf{b})} \exp\left[ \sum_{k=1}^{K} a_k \cos (k \varphi) + b_k \sin(k \varphi) \right]. \label{eq:CircFilt - p multimodal example}
	\end{align}
	The case $K=1$ recovers the previously used von Mises distribution \eqref{eq:CircFilt - vonMises density}, but with a different parametrization.
	For $ K=2 $, this density is referred to as the generalized von Mises distribution, whose properties have been studied extensively \citep{gatto_generalized_2007, gatto_computational_2008}.
	As a proof of concept, we will now use the projection filter for the natural parameters (Eq.~\eqref{eq:ProFilter - Natural Parameters}) to project the solution to the generative model in Eqs.~\eqref{eq:CircFilt - dphi example} and \eqref{eq:CircFilt - dY example} onto a distribution with density \eqref{eq:CircFilt - p multimodal example}.
	
	The circular distributions with densities \eqref{eq:CircFilt - p multimodal example} belong to the exponential family distributions, with natural parameter vector $ \boldsymbol{\theta} = (\mathbf{a},\mathbf{b}) $ with $ \mathbf{a} = \{a_1,\dots,a_K \} $ and $ \mathbf{b} = \{b_1,\dots,b_K \} $, and sufficient statistics given by
	\begin{align}
		T_k^{cos}(\varphi) = \cos (k \varphi), \ \ \ \ T_k^{sin}(\varphi) = \sin (k\varphi).
	\end{align}
	The corresponding expectation parameters are defined by
	\begin{align}
		\eta_k^{cos}= \expecttwo{T_k^{cos}}{\theta} , \ \ \ \ \eta_k^{sin}= \expecttwo{T_k^{sin}}{\theta}.
	\end{align}
	According to Corollary \ref{corollary: Projection filter for exponential families}, the projection filter for the natural parameters of an exponential family density requires us to apply the right-hand side of the gKSE \eqref{eq:CircFilt - KSE circular simple} to the sufficient statistics.
	For this, we need to compute
	\begin{align}
		\expecttwo{\frac{\partial}{\partial \varphi} T_k^{cos}(\varphi) }{\theta} & = - k \expecttwo{ \sin (kx)}{\theta} = -k\, \eta_k^{sin}, \\
		\expecttwo{\frac{\partial^2}{\partial \varphi^2}T_k^{cos}(\varphi) }{\theta} & = - k^2 \expecttwo{ \cos (kx)}{\theta} = - k^2 \, \eta_k^{cos}, \\ 
		\expecttwo{\frac{\partial}{\partial \varphi} T_k^{sin}(\varphi) }{\theta} & = k\, \eta_k^{cos}, \\
		\expecttwo{\frac{\partial^2}{\partial \varphi^2}T_k^{sin}(\varphi) }{\theta} & = - k^2 \, \eta_k^{sin}.
	\end{align}
	
	Furthermore, we note that the components of the Fisher matrix $g_{ij} $ are given by
	\begin{align}
		g_{ij} = & \frac{\partial^2}{\partial \theta_i \partial \theta_j} \log Z(\mathbf{a},\mathbf{b}),
	\end{align}
	where the $ \theta_i $ refer to the $i$th element in the parameter vector $\boldsymbol{\theta}$ that contains the elements of both $\mathbf{a}$ and $\mathbf{b}$.
	Inverting this matrix to get the inverse components $ g^{ij} $ is in general not straightforward.
	In fact, already the Fisher metric needs to be computed numerically, as the normalization $ Z(\mathbf{a},\mathbf{b}) $ is inaccessible in closed form.
	We will thus treat $ G^{-1} $ symbolically and sort the parameters such that $ G^{-1} $ is composed of the blocks
	\begin{align}
		G^{-1} & = \left(\begin{matrix}
			\tilde{G}^{c,c} & \tilde{G}^{s,c} \\\tilde{G}^{s,c} & \tilde{G}^{s,s}
		\end{matrix}\right),
	\end{align}
	
	Then, the projection filter for the natural parameters can be formally written as
	\small
	\begin{align}
	\begin{split}
		d \mathbf{a} & = \tilde{G}^{c,c} \left( - \frac{1}{\kappa_\varphi+\kappa_u} \mathbf{k}^2 \odot \boldsymbol{\eta}^{cos}  dt - \frac{\kappa_u}{\kappa_\varphi+\kappa_u} \mathbf{k} \odot \boldsymbol{\eta}^{sin} \circ d \mathbf{U}_t \right)  \\
		& + \tilde{G}^{s,c} \left( - \frac{1}{\kappa_\varphi+\kappa_u} \mathbf{k}^2 \odot \boldsymbol{\eta}^{sin} \, dt + \frac{\kappa_u}{\kappa_\varphi+\kappa_u} \mathbf{k} \odot \boldsymbol{\eta}^{cos} \circ d \mathbf{U}_t \right), 
	\end{split}\\
	\begin{split}
		d \mathbf{b} & = \tilde{G}^{s,c} \left( - \frac{1}{\kappa_\varphi+\kappa_u} \mathbf{k}^2 \odot \boldsymbol{\eta}^{cos} \, dt - \frac{\kappa_u}{\kappa_\varphi+\kappa_u} \mathbf{k} \odot \boldsymbol{\eta}^{sin} \circ d \mathbf{U}_t \right)  \\
		& + \tilde{G}^{s,s} \left( - \frac{1}{\kappa_\varphi+\kappa_u} \mathbf{k}^2 \odot \boldsymbol{\eta}^{sin} \, dt + \frac{\kappa_u}{\kappa_\varphi+\kappa_u} \mathbf{k} \odot \boldsymbol{\eta}^{cos} \circ d \mathbf{U}_t \right),
	\end{split}
	\end{align}
	\normalsize
	where we denote with $ \mathbf{k} = \left( 1, \dots  , k\right)^\top$ the vector of values $k$, $\mathbf{k}^2$ results from the element-wise squaring of $\mathbf{k}$, $ \boldsymbol{\eta} $ is the vector of expectation parameters, and $ \odot $ the element-wise (Hadamard) product.
	This example demonstrates that our framework could, in principle, be applied project the posterior to more general and inevitably more complicated densities, should the need arise.
	Although we do not show this here explicitly, an instance where this might increase filtering accuracy is one where the initial density $ p_0(\varphi) $ is multimodal.
	The example also shows that, in general, a projection filtering approach might not be the most practical approach.
	Here in particular, computation of the Fisher matrix components might only be possible numerically, and in that case is computationally expensive.
	This highlights the need for a careful choice of the posterior density and its corresponding parametrization, where (ideally) expectations of sufficient statistics are available in closed-form.
\end{example}

\subsection{Quasi continuous-time von Mises observations}
\label{sec:Quasi continuous-time von Mises observations}

So far we have focused on filtering algorithms that rely exclusively on observed angular increments.
Such algorithms are bound to accumulate noise, such that their precision will decay to zero in the long run.
To counteract this effect, let us now consider how to additionally include observations that are generated directly from the hidden state, rather than only its increments.
Specifically, we will in this section propose an observation model for (quasi-)continuous time von Mises valued observations with $ Z_t \in S^1 $, which we will refer to as direct angular observations, because they are governed by the hidden state $ \varphi_t $ directly.
This model will allow us to formulate a von Mises projection filter for both angular observations, and angular increment observations in the continuous-time circular filtering problem.

In classical continuous-time filtering settings, continuous-time observations $ Y_t $ are usually considered to follow a Gaussian diffusion process, whose drift component is governed by the hidden state $ X_t $ \citep{bain_fundamentals_2009}.
Equivalently, one could consider 'time-discretized' (or 'quasi-continuous') observations $ \tilde{Z}_t := \frac{d\tilde{Y}_t}{\Delta t} $ with sampling time step $ \Delta t $, according to
\begin{align}
	\tilde{Z}_t & \sim  \mathcal{N} \left( h(X_t), \sigma_z^2 \Delta t^{-1} \right), \label{eq:von Mises observations - tilde Z_t}
\end{align}
which is the usual setting of discrete-time filtering (with fixed $ \Delta t $), with $ h(x) $ being a potentially nonlinear function.
Notably, the Fisher information $ \mathcal{I}(X_t) $ about the state of the hidden variable $ X_t $ that is conveyed by these quasi-continuous observations $ \tilde{Z}_t $ grows linearly with sampling time step $ \Delta t $ (see Proposition \ref{proposition:Info Scaling Gauss} in Appendix \ref{sec:Apx - info scaling}).
The consequence of this scaling is rather intuitive: decreasing the sampling time step $ \Delta t $ will result in overall more observations per unit time, which, in turn, are individually less informative about the state $ X_t $.
This renders the information rate (information per unit time) independent of the chosen time step.

Analogously, we now consider observations that are drawn from a von Mises distribution centered around a nonlinear $ S^1 $-valued transformation $ h:S^1 \to S^1 $ of the true hidden state $ \varphi_t $,
\begin{align}
	Z_t & \sim \mathcal{VM} ( h(\varphi_t),\alpha (\kappa_z, \Delta t)  ).  \label{eq:von Mises observations - Z_t}
\end{align}
We would like this observation model to have the same linear information scaling properties as the Gaussian observations encountered in the classical filtering problems, i.e.~when hidden state noise and observation noise are uncorrelated.
Thus, we need to choose the function $\alpha (\kappa_z, \Delta t) $ such that the information content about the state $ \varphi_t $ scales linearly with step size $ \Delta t $ and observation precision $ \kappa_z $.

\begin{theorem}
	\label{theorem: kappa scaling}
	If $ \alpha (\kappa_z, \Delta t) $ is chosen such that 
	\begin{align}
		\alpha (\kappa_z, \Delta t)   = \xi^{-1} (\kappa_z \Delta t), \label{eq:von Mises observations - fmin1(Delta t)}
	\end{align}
	where $ \xi^{-1} $ is the inverse of $ \xi(x) = x F(x) $ (and $ F(x) = \frac{I_1(x)}{I_0(x)} $ as defined earlier), then the information about the state of the random variable $ \varphi_t $ scales linearly with sampling time step and observation precision, i.e.,~$ \mathcal{I} (\varphi_t) \propto \kappa_z \, \Delta t$.
\end{theorem}

\begin{figure}
	\centering
	\includegraphics[width=0.9\linewidth]{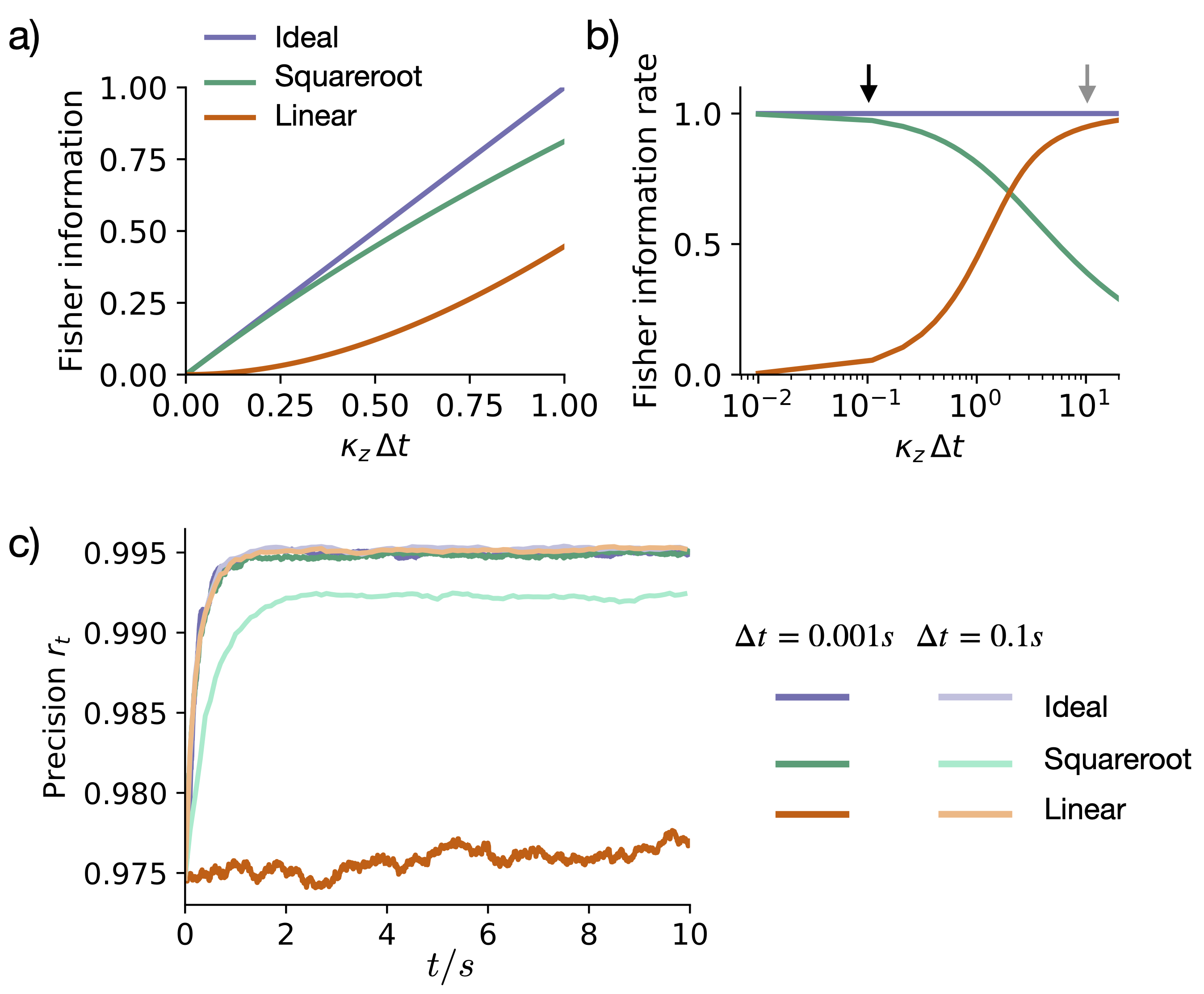}
	\caption{
		{\bf Time scaling of quasi-continuous angular observations.}
		Both a) the Fisher information per observation and b) the Fisher information rate are linear and constant, respectively, in the sampling time step $ \Delta t $ when using $ \alpha(\kappa_z, \Delta t) = \xi^{-1} (\kappa_z \Delta t) $  ("Ideal").
		For comparison, we also plot $ \alpha(\kappa_z, \Delta t) = \sqrt{\kappa_z \Delta t} $ (small $ \kappa_z \Delta t $ approximation, "Squareroot") and  $ \alpha(\kappa_z, \Delta t) = \kappa_z \Delta t $ (Gaussian approximation, "Linear").
		c) Sample simulation with the constant position of the state $ \varphi_t = \varphi $ estimated from quasi-continuous observations with different $ \alpha $ functions and sampling time steps $ \Delta t $.
		Ideally, the estimated precision $ r_t $ should be independent of the chosen simulation time step $ \Delta t $.
		This is satisfied by all simulations except for the linear approximation for small $\kappa_z \Delta t$ (dark orange), and the square root approximation for large $\kappa_z \Delta t$ (light green).
		In these simulations, we used time units of seconds (s), and set $ \kappa_\varphi = 100/s $ and $  \kappa_z = 100/s $ (by design, $\kappa_z$ has units of Fisher information per unit time), without loss of generality.
		Precision estimates were averaged over 10 simulation runs.
		Black and grey arrow in panel b) correspond to the two time step sizes shown in panel c).
	}
	\label{fig:dt scaling}
\end{figure}

\begin{proof}
	The information content about the random variable $ \varphi_t $ is given by the Fisher information
\begin{align}
	\mathcal{I}(\varphi_t) & = \mathbb{E}_{Z_t} \left[ \left(\frac{\partial}{\partial \phi} \log \mathcal{VM} (Z_t ; h(\phi), \alpha  ) \right)^2 | \phi = \varphi_t \right] \\
	&= h'(\varphi_t)^2  \alpha \frac{I_1 (\alpha)}{I_0 (\alpha)} \propto \alpha F(\alpha).
\end{align}
We require that the information content per time step $ \Delta t $ is constant and proportional in $ \kappa_z $, which can be achieved if $ \alpha $ varies with $ \Delta t $ and $ \kappa_z $ according to
\begin{align}
	\alpha (\kappa_z, \Delta t) F(\alpha (\kappa_z, \Delta t) ) & \propto \kappa_z \, \Delta t.
\end{align}
This can be achieved if $ \alpha (\kappa_z, \Delta t) = \xi^{-1} (\kappa_z \, \Delta t) $, with 
\begin{align}
	\xi(x) : = x \cdot \frac{I_1(x)}{I_0(x)}.
\end{align}
\end{proof}
Once a step size $ \Delta t $ is chosen, $ \xi^{-1}(\kappa_z \Delta t) $ can be computed numerically.
For sufficiently small $ \kappa_z \Delta t $, e.g., in the continuum limit, this function can be approximated by $ \xi^{-1}(\kappa_z \Delta t) \approx \sqrt{2 \kappa_z \Delta t} $, while it becomes $ \xi^{-1}(\kappa_z \Delta t) \approx \kappa_z \Delta t$ for large $ \kappa_z \Delta t $ (Figure \ref{fig:dt scaling}).
The latter is consistent with the intuition that, for highly informative observations, the single observation likelihood is well approximated by a Gaussian (which breaks down in the limit $ \Delta t \to 0 $).

What we have considered here is in essence a modified discrete time observation model, which implies that we can take advantage of filtering methods available for circular filtering with discrete-time observations \citep{azmani_recursive_2009,kurz_recursive_2016}.
However, by allowing the precision $ \alpha(\kappa_z,\Delta t) $ to vary with time step, we additionally ensure that the information rate stays constant:
increasing the time step will result in \emph{more} observations per unit time, which is accounted for by less informative individual observations.
Thus, the observation model defined in \eqref{eq:von Mises observations - Z_t} and \eqref{eq:von Mises observations - fmin1(Delta t)} constitutes a \emph{quasi} continuous-time observation model.

\subsection{Adding quasi continuous-time observations to the circular projection filter}

If the measurement function $ h(\varphi) $ in \eqref{eq:von Mises observations - Z_t} is the identity, we can add the direct observations to our filter by straightforwardly making use of Bayes' theorem at every time step.
Specifically, since we assumed our approximated (projected) density to be von Mises at all times, the measurement likelihood
\begin{align}
	p(Z_t|\varphi_t) & = \mathcal{VM} ( Z_t; \varphi_t , \alpha(\kappa_z,\Delta t) ) \label{eq:CircFilt - p(Z phi) linear}
\end{align}
is conjugate to the density before the update $ p_{t^-}(\varphi) := p(\varphi_t = \varphi | Z_{0:t-\Delta T}) = \mathcal{VM} ( \varphi; \mu_{t-}, \kappa_{t-}) $.
In other words, the posterior $ p_t(\varphi) :=  p(\varphi_t = \varphi | Z_{0:t}) $ is guaranteed to be a von Mises density as well:

\small
\begin{align}
	&p_t(\varphi)   \propto   p(Z_t|\varphi_{t} ) \,  p_{t^-}(\varphi)   \\
	&  = \exp \left[  \rowvector{\cos \phi_t}{\sin \phi_t}^\top  \left( \alpha(\kappa_z, dt) \rowvector{\cos Z_t}{\sin Z_t} + \kappa_{t-} \rowvector{\cos \mu_{t-}}{\sin \mu_{t-}} \right)  \right].
\end{align}
\normalsize

\begin{figure*}
	\centering
	\includegraphics[width=0.9\linewidth]{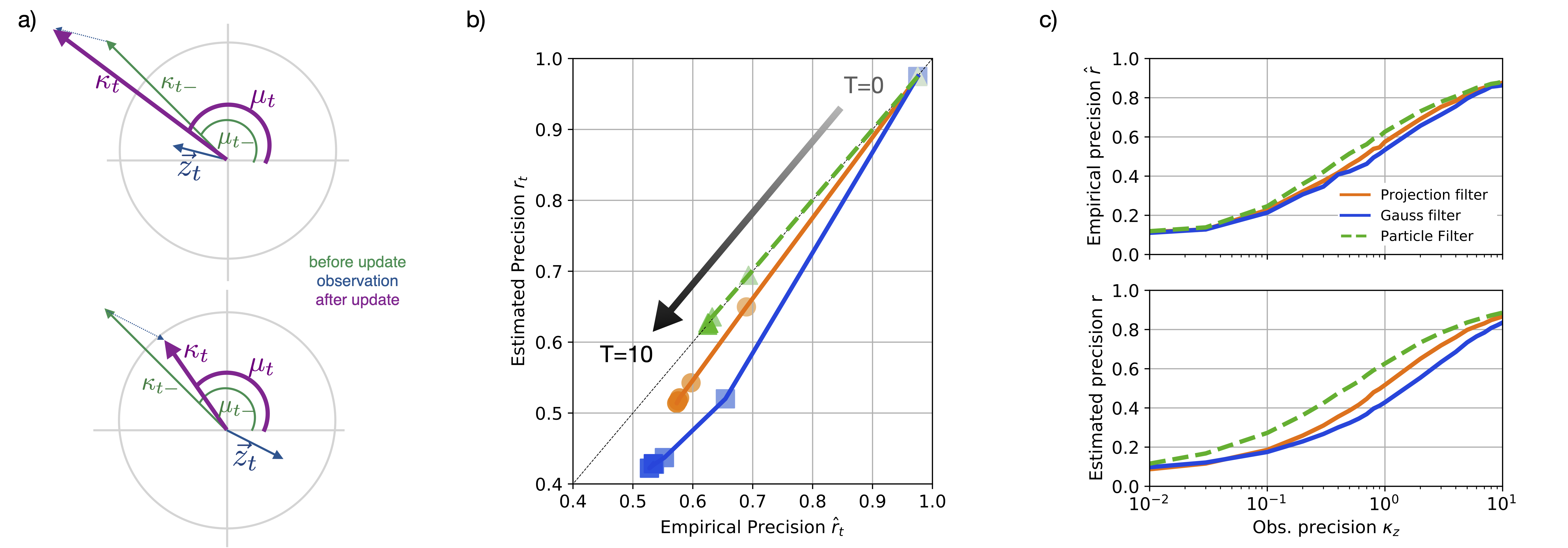}
	\caption{
		{\bf Filtering with quasi-continuous time angular observations.}
		a) Single quasi-continuous time update step \eqref{eq:update direct heading observation} with angular observation $ z_t $, where the length of the vector indicates observation reliability $ \alpha(\kappa_z dt) $. 
		The update step for Bayesian inference on the circle is equivalent to a vector addition in the 2D plane. 
		The lower panel demonstrates that a conflicting observation leads to a decreased certainty of the estimate directly after the update, corresponding to a shorter vector.
		b) Empirical precision $ \hat{r}_t $ (upper panel) and estimated precision $ r_t $ of the circular Kalman filter (circKF) and a Gaussian projection filter (Gauss filter), when compared to a particle filter, for different values of the observation precision $ \kappa_z $ at time $ T=10 \ \kappa_\varphi^{-1} $. 
		Parameters were $ \kappa_\varphi = 1 $ and $ \kappa_u = 1 $, times are in units of $\kappa_\varphi^{-1}$.
		c) Estimated versus empirical precision up to $ T=10 \ \kappa_\varphi^{-1} $ for the different filters at $ \kappa_z = 10 $.
		The precisions shown in b) and c) are averages across 5000 simulation runs.
	}
	\label{fig:direct-observations}
\end{figure*}

As expected from an exponential family distribution, the natural parameters $ \boldsymbol{\theta}_t = \kappa_t (\cos \mu_t , \sin \mu_t)^\top $ are updated according to
\begin{align}
	\boldsymbol{\theta}_t & =  \boldsymbol{\theta}_{t-} + \alpha(\kappa_z,\Delta t)  \rowvector{\cos Z_t}{\sin Z_t}. \label{eq:update direct heading observation}
\end{align}
This operation is equivalent to a summation of vectors in $ \mathbb{R}^2 $, where the natural parameters refer to Eucledian coordinates, and $(\mu, \kappa)$ are the corresponding polar coordinates (Figure \ref{fig:direct-observations}a).
In the continuum limit, we write
\begin{align}
	d \boldsymbol{\theta}_t & = \sqrt{2 \kappa_z dt} \rowvector{\cos Z_t}{\sin Z_t}, \label{eq:dtheta direct heading observation}
\end{align}
where we used that $ \alpha(\kappa_z, dt) \to \sqrt{2 \kappa_z dt} $ for $ dt \to 0 $.
A coordinate transform from $\boldsymbol{\theta}$ to $(\mu, \kappa)$ recovers the update equations for mean $ \mu_t $ and $ \kappa_t $ that result from quasi-continuous time observations
\begin{align}
	d \mu_t & = d \arctan \left(\theta_{2},\theta_{1}\right) = \frac{ \sqrt{2 \kappa_z dt} }{\kappa_t} \sin (Z_t - \mu_t) , \label{eq:CircFilt - dmu - update}\\
	d \kappa_t & = d \sqrt{\theta_{1}^2+\theta_{2}^2} = \sqrt{2 \kappa_z dt} \cos (Z_t - \mu_t). \label{eq:CircFilt - dkappa - update}
\end{align}

The recovered update equations are appealingly simply for identity (or linear) observation functions, as such functions allow us to leverage Bayesian conjugacy properties.
For nonlinear observation functions, in contrast, the posterior after the update is in general \emph{not} in the same class of densities (and in this particular case not a von Mises distribution anymore).
To apply the projection filtering framework to project such a nonlinear observation model back onto the desired manifold of densities, one would need to know the vector field for the density $ p_t $ that incorporates the observation-induced update, a derivation that is beyond the scope of this work.\footnote{As outlined earlier, for a Gaussian setting this would be the update part of the Kushner--Stratonovich equation. To the best of our knowledge, no such equation exists for von Mises-valued observations.}

\begin{example}[The circular Kalman filter]
	\label{sec:Example - circKF}
	Let us revisit angular path integration, which we introduced in Example \ref{sec:Example - circular diffusion increment obs} as a circular diffusion with observed angular increments, and extend it to include direct angular observations $ Z_t $ with likelihood \eqref{eq:CircFilt - p(Z phi) linear}.
	Such angular observations could, for instance, correspond to directly accessible angular cues, such as visual landmarks.
	Combining Eqs.~\eqref{eq:CircFilt - dmu} and \eqref{eq:CircFilt - dkappa} with Eqs.~\eqref{eq:CircFilt - dmu - update} and \eqref{eq:CircFilt - dkappa - update} to include the quasi-continuous updates results in
	\begin{align}
		d \mu_t = &  \frac{\kappa_u}{\kappa_\varphi+\kappa_u}  \cdot d U_t + \frac{ \sqrt{2 \kappa_z dt} }{\kappa_t} \sin (Z_t - \mu_t), \label{eq:CircFilt - dmu - total}\\
		d \kappa_t = &- \frac{1}{2(\kappa_\varphi+\kappa_u)}  \mathcal{F}(\kappa_t) \, dt + \sqrt{2 \kappa_z dt} \,  \cos (\mu_t - Z_t). \label{eq:CircFilt - dkappa - total }
	\end{align}
	Due to the simplicity of these equations, as well as their structural similarity to the generalized Kalman filter, while taking full account of the circular state and measurement space, we coin the SDEs \eqref{eq:CircFilt - dmu - total} and \eqref{eq:CircFilt - dkappa - total } the \emph{circular Kalman filter (circKF)}.
	
	Considering that $\sqrt{2 \kappa_z dt}$ is modulated by the direct observation's reliability $\kappa_z$, the final terms in the filter's update equations nicely reflect the reliability-weighting that is already present in classical filtering problems:
	more reliable direct observations, relative to the current certainty $ \kappa_t $, have a stronger impact on the mean $ \mu_t $.
	Furthermore, if the current observation $ Z_t $ is similar to the current estimate $ \mu_t $, this estimate is hardly updated, while the estimate's certainty $ \kappa_t $ increases.
	In contrast, if direct observations and current estimate are in conflict, the certainty $ \kappa_t $ might temporarily decrease (Figure \ref{fig:direct-observations}a).
	What makes this last feature particularly interesting is that it only occurs in the circKF, but not in its Euclidean counterpart, the standard Kalman filter.
	In the latter an update induced by direct observations \emph{always} leads to an increase in certainty.
	Numerically, the circKF features performance close to that of a particle filter over a wide range of parameters, while outperforming a Gaussian projection filter (Figure \ref{fig:direct-observations}b,c).
	The reason for the deviation of the circKF from the optimal solution is that the von Mises distribution is still an \emph{approximation} of the true posterior, which leads to slight deviations in the updates when the direct observations are integrated.
	These deviations are offset by the more than 10-fold decrease in computation time of the circKF that we observed in simulations:
	a single run in Figure \ref{fig:direct-observations}c with the particle filter took $3.14 \pm 0.11$s, while it only took $0.113 \pm 0.005$s with the circKF on a MacBook Pro (Mid 2019) running 2.3 GHz 8-core Intel Core i9 using NumPy 1.19.2 on Python 3.9.1.
\end{example}

\section{Conclusion}

In this paper, we derived a continuous-time nonlinear filter with observed state increments, based on a projection filtering approach.
Using this framework, we revisited the problem of probabilistic angular path integration.
By additionally proposing a quasi continuous-time model for von Mises-valued direct observations, we were able to formulate a circular filtering algorithm that accounts for both increment and direct angular observations.
Notably, this algorithm fulfills the following four conditions: 
(i) it operates on a circular state-space and 
(ii) in continuous time, 
(iii) it maintains a consistent representation during state propagation and observation update, as ensured by the projection filtering method, and
(iv) it performs the proper integration of both increment and direct observations.
Even though we have only fully worked out the algorithm for univariate circular filtering problems, we have formulated the overall projection filtering framework for more general multivariate problems.
As by the results in \cite{tronarp_continuous-discrete_2020},
we expect our framework to carry over to multivariate circular filtering problems, such as reference vector tracking on the unit sphere.

A possible shortcoming of this approach is that the class of generative models it can deal with is fairly limited.
The generalized Kushner--Stratonovich equation is only valid if the error covariance of the observation process does not explicitly depend on the value of the hidden state.
This constrains the generative model in the following way:
first, we can only allow additive noise in the state process, as any multiplicative noise would enter the increment observation process $\mathbf{U}_t$ as state-dependent noise.
Second, only \emph{linear} transformations of state increments can be considered.
As demonstrated in Example \ref{sec:example Higher-order circular distributions}, another shortcoming of the projection-filtering method in general is that this approach is only computationally feasible for problems that are analytically accessible, i.e., where expectations under the projected density can be computed rapidly, or in closed form.
In this paper, we thus focused on projected densities where these expectations could be efficiently computed.

Despite these limitations, the analytical accessibility and interpretability of our main result, in particular the circular Kalman filter, make it an attractive algorithm for unimodal circular filtering problems.
First, it is straightforward to implement in software.
Since its representation stays fixed through all times, it relies on only two equations which can be integrated straightforwardly, e.g.~with an Euler--Maruyama scheme \cite{kloedenNumericalSolutionStochastic2010}.
As we have seen in our numerical experiments, this makes this filter much faster than established methods, such as particle filters.
Second, the interpretation of the dynamics are intuitively comprehensible.
Third, since it is a continuous-time formulation, it automatically scales with respect to chosen sampling step size (as long as it is sufficiently small).
This is an advantage over continuous-discrete filtering problems, which usually consider a fixed sampling step size, and need to be reformulated should the sampling rate in the observations change or vary across time.
Lastly, animals navigate the world based on a continuous stream of sensory information, which motivates the use of continuous-time models when trying to understand how the brain operates under uncertainty.
Thus, one possible application could be a conceptual description of how the brain performs angular path integration \cite{kutschireiterBayesianPerspectiveRing2021}.

\section*{Appendix}

\subsection{Nonlinear filtering in a nutshell}
\label{sec:Appendix - standard filtering}

Let us briefly review the classical nonlinear filtering setting, i.e.,~filtering with observations that follow a diffusion process with noise uncorrelated to that of the hidden state process, in order to compare this with filtering with observed state increments.
In line with standard literature \cite{bain_fundamentals_2009}, let $\mathbf{Y}_t$ denote the process of $ \mathbb{R}^M $-valued direct observations.
Then the generative model commonly referred to in classical nonlinear filtering is given by
\begin{align}
	d \mathbf{X}_t & = \mathbf{f}(\mathbf{X}_t,t) \, dt + \Sigma_x^{1/2} \, d\mathbf{W}_t, \\
	d\mathbf{Y}_t & = \mathbf{h}(\mathbf{X}_t,t) \, dt + \Sigma_y^{1/2} \, d\tilde{\mathbf{V}}_t, \label{eq:Apx - Standard filtering - dY}
\end{align}
with $\tilde{\mathbf{V}}_t $ and $ \mathbf{W}_t $ independent standard Brownian motion processes, and $ \mathbf{h}: \mathbb{R}^N \times \mathbb{R} \to \mathbb{R}^M$ a potentially nonlinear, vector-valued observation function.
All other quantities have been defined below \eqref{eq:Setting - dX}.
Note that the observations process is a diffusion with error covariance  $ \Sigma_y \, dt $.
The goal of nonlinear filtering is to compute dynamics of the posterior expectations $ \expect{\phi_t} : = \E[\phi(\mathbf{X}_t) | Y_{0:t} ] $, where $ Y_{0:t} = \ \{\mathbf{Y}_\tau:\tau<t\} $ denotes the filtration generated by the process $ \mathbf{Y}_t $.
Formally, this is solved by the Kushner--Stratonovich equation (KSE, \citep[Theorem 3.30]{bain_fundamentals_2009}):
\begin{align}
	d \expect{\phi_t}  = \expect{ \mathcal{L}[\phi_t] } \, dt + &  \cov{\phi_t,\mathbf{h}_t}^T \Sigma_y^{-1}   \left( d\mathbf{Y}_t -  \expect{\mathbf{h}_t} \, dt \right), \label{eq:Apx - KSE}
\end{align}
with $  \cov{\phi_t,\mathbf{h}_t} = \expect{\phi_t , \mathbf{h}_t} - \expect{\phi_t} \expect{\mathbf{h}_t} $.

\subsection{Derivation of the generalized Kushner--Stratonovich equation for the posterior expectation \texorpdfstring{(Eqs.~\eqref{eq:Filtering - gKSE} and \eqref{eq:ProFilter - gKSE in Stratonovich})}{} }
\label{sec:Appendix - derivation gKSE}

Let us revisit the generative model in Eqs.~\eqref{eq:Setting - dX} and \eqref{eq:Setting - dY}:
\begin{align}
	d \mathbf{X}_t & = \mathbf{f}(\mathbf{X}_t,t) \, dt + \Sigma_x^{1/2} \, d\mathbf{W}_t, \label{eq:Setting - dX - SI}\\
	d \mathbf{U}_t & = C \, \mathbf{f}(\mathbf{X}_t,t) \, dt + C \Sigma_x^{1/2} \, d\mathbf{W}_t + \Sigma_u^{1/2} \, d\mathbf{V}_t, \label{eq:Apx - dU}
\end{align}
with $ \mathbf{W}_t $ and $ \mathbf{V}_t $ independent vector-valued Brownian motion processes, as defined earlier.
Here, we derive the gKSE \eqref{eq:Filtering - gKSE} by treating this model as a correlated noise filtering problem, which allows us to directly apply established results \cite[Corollary 3.38]{bain_fundamentals_2009}.

\begin{lemma}
	\label{lemma: gKSE}
	The generalized Kushner--Stratonovich equation (gKSE) for the evolution of the conditional expectation of a test function $ \expect{\phi_t} = \mathbb{E}\left[ \phi(X_t) | Y_{0:t} \right] $ in the presence of state increment observations $ d \mathbf{U}_t $ is given by (It\^{o} form)
	\begin{align}
	\begin{split}
		d \expect{\phi_t}  = \expect{ \mathcal{L}[\phi_t] } dt + &   \left( \cov{\phi_t, \mathbf{f}_t}  + \Sigma_x\expect{ \nabla \phi_t} \right)^T   \\
		& \cdot C^T \tilde{\Sigma}_u^{-1} (d\mathbf{U}_t - C \expect{\mathbf{f}_t} \, dt).  \label{eq:Apx - gKSE}
	\end{split}
	\end{align}
	
\end{lemma}

\begin{proof}
	To simplify calculations, we first require that the observations process has an error covariance that equals the identity, and thus we rescale the process $ d \mathbf{U}_t $ by $ \tilde{\Sigma}_u^{1/2} =  (C \Sigma_x C^T + \Sigma_u)^{1/2} $,
	\begin{align}
		d \tilde{\mathbf{U}}_t = \tilde{\Sigma}_u^{-1/2} d \mathbf{U}_t. \label{eq:Apx = dYtilde} 
	\end{align}
	Eqs.~\eqref{eq:Setting - dX - SI} and \eqref{eq:Apx = dYtilde} define a filtering problem where the noise in hidden process $ \mathbf{X}_t $ and observation process $ \tilde{\mathbf{U}}_t $ are correlated, with quadratic covariation
	\begin{align}
		[\mathbf{X},\tilde{\mathbf{U}}^T]_t = \Sigma_x C^T \tilde{\Sigma}_u^{-1/2}.
	\end{align}
	Thus, the result of \citep[Corollary 3.38]{bain_fundamentals_2009} (KSE for correlated noise filtering problems) is directly applicable to our problem.
	For more details on \citep[Corollary 3.38]{bain_fundamentals_2009}, we kindly refer to the proof based on the innovations method provided therein.
	An alternative proof based on the change of measure approach is presented in \citep{nusken_state_2019}.
	
	Note that the KSE for correlated noise \citep[Eq.~3.72]{bain_fundamentals_2009} is similar to the classical KSE for uncorrelated noise problems \eqref{eq:Apx - KSE}, except for a correction term, given by a vector field $ \mathcal{B}[\phi] $,
	\begin{align}
	\begin{split}
		d \expect{\phi_t}  = \expect{ \mathcal{L}[\phi_t] } \, dt + & \left(  \cov{\phi_t, \tilde{\Sigma}_u^{-1/2} C \mathbf{f}_t} + \expect{\mathcal{B} [\phi_t]}  \right)^T  \\
		& \cdot (d\tilde{\mathbf{U}} - \tilde{\Sigma}_u C \expect{\mathbf{f}_t} dt).
	\end{split}
	\end{align}
	The vector field $ \mathcal{B}[\phi] $ can be read out from the dynamics of the quadratic covariation between the process $ \phi_t $ and the rescaled observations process $ \tilde{\mathbf{U}}_t $.
	Using It\^{o}'s lemma, we can write:
	\begin{align}
		d \phi_t & =( \nabla \phi_t)^T \cdot d\mathbf{X}_t + \frac{1}{2} \text{Tr}\left( \Sigma_x H_\phi \right) \, dt,
	\end{align}
	and thus
	\begin{align}
		d [\phi, \tilde{\mathbf{U}}^T ]_t & = ( \nabla \phi_t)^T d[\mathbf{X},\tilde{\mathbf{U}}^T ]_t  \\
		& = ( \nabla \phi_t)^T  \Sigma_x C^T \tilde{\Sigma}_u^{-1/2}  dt =: \mathcal{B}[\phi_t]^T dt.
	\end{align}
	Further identifying $ \mathbf{h}_t = \tilde{\Sigma}_u^{-1/2} C\, \mathbf{f}_t $, we find
	\begin{align}
	\begin{split}
		d \expect{\phi_t}  = & \expect{ \mathcal{L}[\phi_t] } \, dt +    \left( \cov{\phi_t, \mathbf{f}_t} + \Sigma_x \expect{\nabla \phi_t }\right)^T  \\
		& \quad \quad \cdot C^T \tilde{\Sigma}_u^{-1/2} (d\tilde{\mathbf{U}}_t - \tilde{\Sigma}_u^{-1/2} C \expect{\mathbf{f}_t} \, dt).
	\end{split}
	\end{align}
	Rescaling $ d\tilde{\mathbf{U}}_t = \tilde{\Sigma}_u^{-1/2} d \mathbf{U}_t $ yields \eqref{eq:Apx - gKSE}.
\end{proof}

\begin{remark}
	By combining \eqref{eq:Apx - KSE} and \eqref{eq:Apx - gKSE} it is possible to derive a generalized Kushner equation when both types of observations are present, i.e., when we consider both the process $ \mathbf{Y}_t $ (Eq.~\ref{eq:Apx - Standard filtering - dY}) and the process $ \mathbf{U}_t $ (Eq.~\ref{eq:Apx - dU}) as the observations:
	\begin{align}
	\begin{split}
		d \expect{\phi_t}  = & \expect{ \mathcal{L}[\phi_t] } \, dt +  \cov{\phi_t,\mathbf{h}_t}^T \Sigma_y^{-1}   \left( d\mathbf{Y}_t -  \expect{\mathbf{h}_t} \, dt \right)  \\
		&+  \left( \cov{\phi_t, \mathbf{f}_t} + \Sigma_x \expect{\nabla \phi_t} \right)^T C^T \tilde{\Sigma}_u^{-1/2}  \\
		& \quad  \cdot (d\tilde{\mathbf{U}}_t - \tilde{\Sigma}_u^{-1/2} C \expect{\mathbf{f}_t} \, dt).
	\end{split}
	\end{align}
	In this case, the expectations $ \expect{\cdot} $ are with respect to the filtration $ Y_{0:t} $ and $ U_{0:t} $, i.e., $ \expect{\phi_t} =  \mathbb{E} \left[ \phi(\mathbf{X}_t) | Y_{0:t}, U_{0:t} \right]$.
\end{remark}

\begin{corollary}
	\label{corollary: gKSE Stratonovich}
	In Stratonovich form, the generalized Kushner--Stratonovich equation (gKSE) for the evolution of the conditional expectation of a test function $ \expect{\phi_t} = \mathbb{E}\left[ \phi(X_t) | Y_{0:t} \right] $ in the presence of increment observations $ d \mathbf{U}_t $ reads:
	\begin{align}
	\begin{split}
				 d \expect{\phi_t}  = &   \bigg[ \expect{\tilde{\mathcal{L}}[\phi_t]}  - \frac{1}{2} \bigg(  \cov{\phi_t, ||\tilde{\mathbf{\Sigma}}_u^{-1/2} C \, \mathbf{f}_t ||^2 }  \\
		& \quad + \tr{\tilde{\Sigma}_u^{-1} C \cdot \cov{\phi_t,J_f} \Sigma_x C^T} \bigg) \bigg] \, dt  \\
		&  + \big( \cov{\phi_t, \mathbf{f}_t}  + \Sigma_x\expect{ \nabla \phi_t} \big)^T C^T \tilde{\Sigma}_u^{-1}  \circ d\mathbf{U}_t , \label{eq:Apx - gKSE in Stratonovich}
	\end{split}
	\end{align}
	with
	\begin{align}
	\begin{split}
	\expect{\tilde{\mathcal{L}} [\phi_t]} = & \expect{\mathcal{L}[\phi_t]} - \expect{(\nabla \phi_t)^T \Sigma_x C^T \tilde{\Sigma}_u^{-1} C \, \mathbf{f}_t }  \\
	& -\frac{1}{2} \tr{ \Sigma_x \expect{H_\phi}  \Sigma_x C^T \tilde{\Sigma}_u^{-1} C }
	\end{split}
	\end{align}
	For one-dimensional problems with $ N=M=1 $, $ C = c $, $ \Sigma_x = \sigma_x^2 $, $ \Sigma_u = \sigma_u^2 $ and $ \tilde{\Sigma}_u = \tilde{\sigma}_u^2 = c^2 \sigma_x^2 + \sigma_u^2 $, \eqref{eq:Apx - gKSE in Stratonovich} simplifies to
	\begin{align}
	\begin{split}
	d \expect{\phi_t}  = &  \bigg[  \frac{\sigma_u^2}{\tilde{\sigma}_u^2} \expect{ \mathcal{L}[\phi_t] }  \\
	& \quad  -\frac{c^2}{2 \tilde{\sigma}_u^2} \Big( \cov{\phi_t, f_t^2 }  + \sigma_x^2  \cov{\phi_t, \frac{\partial}{\partial x} f_t }\Big) \bigg] \, dt  \\
	& + \frac{c}{\tilde{\sigma}_u^2}  \bigg[ \cov{\phi_t, f_t}  + \sigma_x^2 \expect{\frac{\partial}{\partial x} \phi_t } \bigg] \circ dU_t . \label{eq:Apx - gKSE in Stratonovich 1D}
	\end{split}
	\end{align}
\end{corollary}

\begin{proof}
	We convert between It\^{o} and Stratonovich calculus using the Wong--Zakai theorem \cite{wong_relation_1965}:
	\begin{align}
		\mathbf{B}_t^T \cdot d \mathbf{U}_t & = \mathbf{B}_t^T \circ d \mathbf{U}_t - \frac{1}{2} d [\mathbf{B}^T,\mathbf{U}]_t,
	\end{align}
	where the symbol $ \circ $ denotes Stratonovich calculus, and $ [B^T,Y]_t $ is the quadratic covariation between the processes $ Y_t $ and $ \mathbf{B}_t $.
	We identify (cf.~Eq.~\eqref{eq:Apx - gKSE})
	\begin{align}
		\mathbf{B}_t^T =   \left( \cov{\phi_t, \mathbf{f}_t} + \Sigma_x \expect{ \nabla \phi_t} \right)^T C^T \tilde{\Sigma}_u^{-1},
	\end{align}
	and write
	\small
	\begin{align}
			& d \expect{\phi_t}  = \expect{ \mathcal{L}[\phi_t] } \, dt - \mathbf{B}_t^T C  \expect{\mathbf{f}_t} \, dt + \mathbf{B}_t^T  \cdot d \mathbf{U}_t  \\
			& \quad =  \expect{ \mathcal{L}[\phi_t] } \, dt - \mathbf{B}_t ^T C \expect{\mathbf{f}_t} \, dt + \mathbf{B}_t^T \circ d \mathbf{U}_t  - \frac{1}{2} d[\mathbf{B}^T,\mathbf{U}]_t. \label{eq:Apx - dphi Strat temp}
	\end{align}
	\normalsize

	To obtain the change in quadratic covariation $ d[\mathbf{B}^T,\mathbf{U}]_t $, it is helpful to find $ d \mathbf{B}_t $ by It\^{o}'s lemma
	\begin{align}
	\begin{split}
		d \mathbf{B}^T_t = \Big( & d \expect{\phi_t \mathbf{f}_t} - \expect{\phi_t} \, d\expect{\mathbf{f}_t} - d \expect{\phi_t} \expect{\mathbf{f}_t}   \\
		& - d \expect{\phi_t} \, d\expect{\mathbf{f}_t} +  d\expect{ \nabla\phi_t } \Sigma_x \Big) C^T \tilde{\Sigma}_u^{-1}.
	\end{split}
	\end{align}
	The evolution of the expectations is obtained by straightforward application of the gKSE \eqref{eq:Apx - gKSE}, substituting $ \phi_t  $ with the functions $ \phi_t \mathbf{f}_t $, $ \mathbf{f}_t$ and $\frac{\partial}{\partial x} \phi_t $.
	This will result in terms that multiply $ d \mathbf{U}_t $, which are those relevant for computing the change of the covariation process, $ d [\mathbf{B}^T,\mathbf{U}]_t $.
	Further note that the quadratic covariation of the observation process evolves according to $ d[\mathbf{U}^T,\mathbf{U}]_t = \tr{\Sigma_u} dt$.
	Some tedious but straightforward algebra and term rearrangements then result for the quadratic covariation in
	\begin{align}
	\begin{split}
		 d [\mathbf{B}^T,\mathbf{U}]_t =  &  \Big[  \cov{\phi_t, || \tilde{\Sigma}_u^{-1/2} C \, \mathbf{f}_t ||^2}  \\
		 & \quad + \tr{C^T \tilde{\Sigma}_u^{-1} C \cdot \cov{\phi_t, J_f} \Sigma_x }   \\
		 & \quad + 2 \cov{ (\Sigma_x C^T \tilde{\Sigma}_u^{-1} C \, \mathbf{f}_t)^T, \nabla \phi_t }  \\
		 & \quad - 2 \cov{\phi_t, \mathbf{f}_t} C^T \tilde{\Sigma}_u^{-1} C \expect{\mathbf{f}_t}^T  \\
		 & \quad + \tr{\Sigma_x \expect{H_\phi} \Sigma_x C^T \tilde{\Sigma}_u^{-1} C } \Big] dt.
	\end{split}
	\end{align}
	Plugging this into \eqref{eq:Apx - dphi Strat temp}, and again some algebra, yields
	\begin{align}
	\begin{split}
		 d \expect{\phi_t} &  =  \Big[ \expect{\mathcal{L}[\phi_t]} - \expect{(\nabla \phi_t)^T \Sigma_x C^T \tilde{\Sigma}_u^{-1} C \, \mathbf{f}_t }  \\
		& \quad -\frac{1}{2} \tr{ \Sigma_x \expect{H_\phi}  \Sigma_x C^T \tilde{\Sigma}_u^{-1} C } \Big] dt  \\
		& - \frac{1}{2} \big[  \cov{\phi_t, ||\tilde{\mathbf{\Sigma}}_u^{-1/2} C \, \mathbf{f}_t ||^2 }  \\
		& \quad + \tr{\tilde{\Sigma}_u^{-1} C \cdot \cov{\phi_t,J_f} \Sigma_x C^T}\Big] \, dt  \\
		&  \left( \cov{\phi_t, \mathbf{f}_t}  + \Sigma_x\expect{ \nabla \phi_t} \right)^T C^T \tilde{\Sigma}_u^{-1}  \circ d\mathbf{U}_t . \label{eq:Apx - dphi Strat}
	\end{split}
	\end{align}
	This equation can be further simplified by noting that $ \mathbb{I} - \Sigma_x C^T \tilde{\Sigma}_u^{-1} C = C^{-1} \Sigma_u \tilde{\Sigma}_u^{-1} C $.
	Further, we can substitute
	\begin{align}
	\begin{split}
		\expect{\tilde{\mathcal{L}} [\phi_t]} = &\expect{\mathcal{L}[\phi_t]} - \expect{(\nabla \phi_t)^T \Sigma_x C^T \tilde{\Sigma}_u^{-1} C \, \mathbf{f}_t }  \\
		  & -\frac{1}{2} \tr{ \Sigma_x \expect{H_\phi}  \Sigma_x C^T \tilde{\Sigma}_u^{-1} C }
	\end{split}
	\end{align}
	in the first line, which yields Eq.~\eqref{eq:Apx - gKSE in Stratonovich}.
	Equation \eqref{eq:Apx - gKSE in Stratonovich 1D} follows from \eqref{eq:Apx - gKSE in Stratonovich} as the 1D special case.
\end{proof}

\subsection{Fisher metric and scalar product}
\label{sec:Apx - Fisher metric}

\begin{proposition}
	\label{proposition: scalar product and Fisher metric}
	The scalar product defined in Eq.~\eqref{eq:ProFilter - scalar product},
	\begin{align}
		\innerproduct{Z_1,Z_2}_\theta := & \int_\Omega d\mathbf{x}\,\frac{Z_1(\mathbf{x}) Z_2 (\mathbf{x})}{p_\theta(\mathbf{x})}, \ \ \ Z_1,Z_2\in T_{\theta}\mathcal{S}, \label{eq:Apx - scalar product}
	\end{align}
	 is the scalar product on $ T_{\theta}\mathcal{S} $ that is associated with the Fisher metric \citep{amari_differential-geometrical_1990}
	 \begin{align}
	 	g_{ij} &=  \E_{p_\theta} \left[ \frac{\partial \log p_\theta(\mathbf{x})}{\partial \theta_i}  \frac{\partial \log p_\theta(\mathbf{x})}{\partial \theta_j} \right]. \label{eq:Apx - Fisher metric}
	 \end{align}
	 
\end{proposition}
\begin{proof}
Consider $ Z_1 $ and $ Z_2 $ to correspond to two of the basis vectors of the tangent space $ T_{\theta}\mathcal{S} $, i.e., $ Z_1 = \frac{\partial p_\theta(\mathbf{x})}{\partial \theta_{i}} $ and $ Z_1 = \frac{\partial p_\theta(\mathbf{x})}{\partial \theta_{j}} $.
Then
\small
\begin{align}
 \innerproduct{\frac{\partial p_\theta}{\partial \theta_{i}},\frac{\partial p_\theta}{\partial \theta_{j}}}_\theta & =  \int_\Omega d\mathbf{x}\,\frac{1}{p_\theta(\mathbf{x})} \frac{\partial p_\theta(\mathbf{x})}{\partial \theta_{i}}  \frac{\partial p_\theta(\mathbf{x})}{\partial \theta_{j}}  \\
	& = \int_\Omega d\mathbf{x}\, \frac{\partial \log p_\theta(\mathbf{x})}{\partial \theta_{i}}  \frac{\partial \log p_\theta(\mathbf{x})}{\partial \theta_{j}} p_\theta(\mathbf{x})  \\
	& = \mathbb{E} \left[ \frac{\partial \log p_\theta(\mathbf{x})}{\partial \theta_{i}}  \frac{\partial \log p_\theta(\mathbf{x})}{\partial \theta_{j}} \right] = g_{ij}.
\end{align}
\normalsize
This concludes the proof.
\end{proof}

\subsection{Information scaling}
\label{sec:Apx - info scaling}

\begin{proposition}
	\label{proposition:Info Scaling Gauss}
	The Fisher information $ \mathcal{I}(X_t) $ about the hidden state variable $ X_t $ that is conveyed by Gaussian-type discrete-time observations, 
	\begin{align}
		\tilde{Z}_t & \sim \mathcal{N} \left( \tilde{Z}_t ; g(X_t), \Delta t^{-1} \right),
	\end{align}
	grows linearly with the time step $ \Delta t $.
\end{proposition}

\begin{proof}
	The information content about the variable $ \varphi_t $ that is conveyed by the observation $ \tilde{Z}_t $ is given by the Fisher information
	\begin{align}
		\mathcal{I}(X_t) & = \mathbb{E}_{\tilde{Z}_t} \left[ \left(\frac{\partial}{\partial x} \log \mathcal{N} (\tilde{Z}_t ; g(x), \frac{\sigma_z^2}{\Delta t}  ) \right)^2 | x = X_t \right] \nonumber \\
		& = \frac{g'(x)^2}{\sigma_z^4} (\Delta t)^2 \, \mathbb{E}_{\tilde{Z}_t} \left[ (\tilde{Z}_t - g(x))^2 \right] \nonumber \\
		& = \frac{g'(x)^2}{\sigma_z^2} \Delta t \, \propto \, \Delta t. \nonumber
	\end{align}
\end{proof}

\subsection{Details on numerical simulations}
\label{sec:Apx - numerics details}
Our numerical simulations in Figures \ref{fig:Increment-Observations} and \ref{fig:direct-observations}, corresponding to Examples \ref{sec:Example - circular diffusion increment obs} and \ref{sec:Example - circKF}, were based on artificial data generated from the true model equations.
In particular, the ``true'' state $\varphi_t$ is a single trajectory from Eq.~\eqref{eq:CircFilt - dphi example}, and observations are drawn at each time point from Eq.~\eqref{eq:CircFilt - dY example} and, in the case of Example \ref{sec:Example - circKF}, additionally from Eq.~\eqref{eq:CircFilt - p(Z phi) linear}.
To simulate trajectories and observations, we use the Euler--Maruyama approximation \cite{kloedenNumericalSolutionStochastic2010}.
In this approximation, the time-discretized generative model with fixed time step size $\Delta t$ in Examples \ref{sec:Example - circular diffusion increment obs} and \ref{sec:Example - circKF} reads:
\begin{align}
	\varphi_t & \sim \mathcal{N} ( \varphi_{t-\Delta t} , \frac{1}{\kappa_\varphi} \Delta t) \mod 2 \pi \\
	\Delta U_t & \sim \mathcal{N} (	 \varphi_t - \varphi_{t-\Delta t} , \frac{1}{\kappa_u} \Delta t) \\
	Z_t & \sim \mathcal{VM}( \varphi_t , \xi^{-1} (\kappa_z \Delta t) ).
\end{align}
The same time-discretization scheme was used to numerically integrate the SDEs \eqref{eq:CircFilt - dmu}, \eqref{eq:CircFilt - dkappa}, \eqref{eq:CircFilt - dmu - total} and \eqref{eq:CircFilt - dkappa - total } for the von Mises parameters $\mu_t$ and $\kappa_t$.
Unless stated otherwise, we used $\Delta t = 0.01$ in all our numerical simulations, and give times in units of $\kappa_\varphi$.

\subsection{Benchmarks for numerical simulations}
\label{sec:Apx - numerics}

For our numerical simulations in Figures \ref{fig:Increment-Observations} and \ref{fig:direct-observations}, corresponding to Examples \ref{sec:Example - circular diffusion increment obs} and \ref{sec:Example - circKF}, we used the following filtering algorithms to compare the circKF against.

\subsubsection{Particle Filter}

As benchmark, we used a Sequential Importance Sampling/Resampling particle filter (SIS-PF, \cite{doucet_sequential_2010}), that was modified to account for state increment observations $ \Delta U_t $.

The $ N $ particles in the SIS-PF where propagated according to
\small 
\begin{align}
\begin{split}
	&\pi(\varphi_t^{(j)} | \varphi_{t-\Delta t}^{(j)} , \Delta U_t)  \\
	&\quad  =  \mathcal{N} ( \varphi_{t-\Delta t}^{(j)} +  \frac{\kappa_u}{\kappa_u + \kappa_\varphi} \Delta U_t , \frac{1}{\kappa_\varphi+\kappa_u} \Delta t) \mod 2 \pi , 
\end{split}
\end{align}
\normalsize
and each particle $ j $ was weighted at each time step according to
\begin{align}
	w_t^{(j)} & = w_{t-\Delta t}^{(j)} \cdot  \mathcal{VM} (Z_t ;  \varphi_t^{(j)} , \xi^{-1} (\kappa_z \Delta t) ),
\end{align}
yielding an SIS for this model that is asymptotically exact in the $N \to \infty$ limit.
Mean $ \mu_t $ and precision $ r_t $ of the filtering distribution were determined at each time step according to a weighted average on the circle, i.e.~the first circular moment:
\begin{align}
	r_t \exp(i \mu_t) & = \sum_{j=1}^N w_t^{(i)} \exp( i \varphi_t^{(i)} ).
\end{align}
\normalsize
For a von Mises distribution, the radius $ r $ of the first circular moment and the precision parameter $ \kappa $ are related via $ r = \frac{I_1(\kappa)}{I_0(\kappa)} $, which is why we use $ r $ rather than $ \kappa $ in our plots.

In our simulations, we used $ N=10^3 $ if direct angular observations $ Z_t $ were present, and $ N=10^4 $ if only state increment observations were present.
We re-sampled the particles whenever the effective number of particles, $ N_\text{eff} = \sum_j (w^{(i)})^{-2} $, was lower than $ N/2 $.

\subsubsection{Gaussian approximation}

The reference filter, which we refer to as ``Gauss filter'', is a heuristic method that assumes posterior mean $ \mu_t $ and variance $ \sigma_t $ to evolve according to a generalized Kalman--Bucy filter (Eqs.~\eqref{eq:gKBF - dmu} and \eqref{eq:gKBF - dsig2}).
Such a filter is often referred to as ``assumed density filter'' (ADF) in the literature, which under certain conditions, such as for the circular filtering problem we consider here, becomes fully equivalent to a Gaussian projection filter (see \citep[Section 7]{brigo_approximate_1999} for in-depth discussion).
In order to make the resulting distribution circular, this Gaussian is consecutively approximated by a von Mises distribution via $ \kappa_t \approx \sigma_t^{-2} $, resulting in the following update equations for the model in Example \ref{sec:Example - circular diffusion increment obs}:
\begin{align}
	d\mu_t & = \frac{\kappa_u}{\kappa_\varphi + \kappa_u} \, d U_t, \\
	d \kappa_t & = d \left( \frac{1}{\sigma_t^2} \right) = - \frac{1}{\kappa_\varphi + \kappa_u } \cdot \frac{1}{\kappa_t^2} \, dt .
\end{align}
For the model used in Example \ref{sec:Example - circKF}, this is combined with the observations in the same way as for the circKF.
Note that in absence of direct angular observations $ z_t $, the mean dynamics are the same as for the circKF, while $ \kappa_t $ deviates (as shown in Figure \ref{fig:Increment-Observations}). 
When direct angular observations are present, this, in turn, affects the computation of the update in the mean dynamics, which leads to a worse numerical performance than the circKF.

\subsection{Code availability}

Jupyter notebooks to generate Figs.~1-3, the underlying simulation data as well as Python scripts to generate this data has been deposited at Zenodo, and is publicly available at \href{https://doi.org/10.5281/zenodo.5820406}{https://doi.org/10.5281/zenodo.5820406}.

\section*{Acknowledgements}
We would like to thank Melanie Basnak and Rachel Wilson (Harvard Medical School) for active and ongoing discussions that motivated the question of probabilistic path integration, and who helped us formulate the questions that we address in this manuscript.
We further would like to thank Johannes Bill (Harvard Medical School) and Simone Surace (University of Bern), as well as the anonymous reviewers, for helpful feedback and suggestions on the manuscript.

\section*{Funding}
A.K. was funded by the Swiss National Science Foundation (grant numbers P2ZHP2\_184213 and P400PB\_199242).
J.D. and L.R. were supported by the Harvard Medical School Dean’s Initiative Award Program for Innovative Grants in the Basic and Social Sciences, and a grant from the National Institutes of Health (NIH/NINDS, 1R34NS123819).

%

\small
\begin{thebibliography}{32}
	\providecommand{\natexlab}[1]{#1}
	\providecommand{\url}[1]{#1}
	\csname url@samestyle\endcsname
	\providecommand{\newblock}{\relax}
	\providecommand{\bibinfo}[2]{#2}
	\providecommand{\BIBentrySTDinterwordspacing}{\spaceskip=0pt\relax}
	\providecommand{\BIBentryALTinterwordstretchfactor}{4}
	\providecommand{\BIBentryALTinterwordspacing}{\spaceskip=\fontdimen2\font plus
		\BIBentryALTinterwordstretchfactor\fontdimen3\font minus
		\fontdimen4\font\relax}
	\providecommand{\BIBforeignlanguage}[2]{{%
			\expandafter\ifx\csname l@#1\endcsname\relax
			\typeout{** WARNING: IEEEtranN.bst: No hyphenation pattern has been}%
			\typeout{** loaded for the language `#1'. Using the pattern for}%
			\typeout{** the default language instead.}%
			\else
			\language=\csname l@#1\endcsname
			\fi
			#2}}
	\providecommand{\BIBdecl}{\relax}
	\BIBdecl
	
	\bibitem[Heinze et~al.(2018)Heinze, Narendra, and
	Cheung]{heinze_principles_2018}
	\BIBentryALTinterwordspacing
	S.~Heinze, A.~Narendra, and A.~Cheung, ``Principles of insect path
	integration,'' \emph{Current biology : CB}, vol.~28, no.~17, pp.
	R1043--R1058, Sep. 2018. [Online]. Available:
	\url{https://www.ncbi.nlm.nih.gov/pmc/articles/PMC6462409/}
	\BIBentrySTDinterwordspacing
	
	\bibitem[Kreiser et~al.(2018)Kreiser, Cartiglia, Martel, Conradt, and
	Sandamirskaya]{kreiser_neuromorphic_2018}
	R.~Kreiser, M.~Cartiglia, J.~N.~P. Martel, J.~Conradt, and Y.~Sandamirskaya,
	``A {Neuromorphic} {Approach} to {Path} {Integration}: {A} {Head}-{Direction}
	{Spiking} {Neural} {Network} with {Vision}-driven {Reset},'' in \emph{2018
		{IEEE} {International} {Symposium} on {Circuits} and {Systems} ({ISCAS})},
	May 2018, pp. 1--5, iSSN: 2379-447X.
	
	\bibitem[Taube et~al.(1990)Taube, Muller, and Ranck]{taube_head-direction_1990}
	\BIBentryALTinterwordspacing
	J.~S. Taube, R.~U. Muller, and J.~B. Ranck, ``Head-direction cells recorded
	from the postsubiculum in freely moving rats. {I}. {Description} and
	quantitative analysis.'' \emph{The Journal of Neuroscience}, vol.~10, no.~2,
	pp. 420--35, 1990. [Online]. Available:
	\url{http://www.ncbi.nlm.nih.gov/pubmed/2303851}
	\BIBentrySTDinterwordspacing
	
	\bibitem[Seelig and Jayaraman(2015)]{seelig_neural_2015}
	J.~D. Seelig and V.~Jayaraman, ``Neural dynamics for landmark orientation and
	angular path integration,'' \emph{Nature}, vol. 521, no. 7551, pp. 186--191,
	2015.
	
	\bibitem[Skaggs et~al.(1995)Skaggs, Knierim, Kudrimoti, and
	McNaughton]{skaggs_model_1995}
	W.~E. Skaggs, J.~J. Knierim, H.~S. Kudrimoti, and B.~L. McNaughton,
	``\BIBforeignlanguage{en}{A {Model} of the {Neural} {Basis} of the {Rat}'s
		{Sense} of {Direction}},'' \emph{\BIBforeignlanguage{en}{Advances in neural
			information processing systems}}, p.~10, 1995.
	
	\bibitem[Turner-Evans et~al.(2017)Turner-Evans, Wegener, Rouault, Franconville,
	Wolff, Seelig, Druckmann, and Jayaraman]{turner-evans_angular_2017}
	\BIBentryALTinterwordspacing
	D.~Turner-Evans, S.~Wegener, H.~Rouault, R.~Franconville, T.~Wolff, J.~D.
	Seelig, S.~Druckmann, and V.~Jayaraman, ``\BIBforeignlanguage{en}{Angular
		velocity integration in a fly heading circuit},''
	\emph{\BIBforeignlanguage{en}{eLife}}, vol.~6, p. e23496, May 2017. [Online].
	Available: \url{https://elifesciences.org/articles/23496}
	\BIBentrySTDinterwordspacing
	
	\bibitem[Xie et~al.(2002)Xie, Hahnloser, and Seung]{xie_double-ring_2002}
	X.~Xie, R.~H. Hahnloser, and H.~S. Seung, ``Double-ring network model of the
	head-direction system,'' \emph{Physical Review E - Statistical Physics,
		Plasmas, Fluids, and Related Interdisciplinary Topics}, vol.~66, no.~4, pp.
	9--9, 2002.
	
	\bibitem[Ma et~al.(2006)Ma, Beck, Latham, and Pouget]{ma_bayesian_2006}
	\BIBentryALTinterwordspacing
	W.~J. Ma, J.~M. Beck, P.~E. Latham, and A.~Pouget, ``Bayesian inference with
	probabilistic population codes.'' \emph{Nature Neuroscience}, vol.~9, no.~11,
	pp. 1432--8, Nov. 2006. [Online]. Available:
	\url{http://www.ncbi.nlm.nih.gov/pubmed/17057707}
	\BIBentrySTDinterwordspacing
	
	\bibitem[Fiser et~al.(2010)Fiser, Berkes, Orbán, and
	Lengyel]{fiser_statistically_2010}
	\BIBentryALTinterwordspacing
	J.~Fiser, P.~Berkes, G.~Orbán, and M.~Lengyel,
	``\BIBforeignlanguage{en}{Statistically optimal perception and learning: from
		behavior to neural representations},'' \emph{\BIBforeignlanguage{en}{Trends
			in Cognitive Sciences}}, vol.~14, no.~3, pp. 119--130, Mar. 2010. [Online].
	Available:
	\url{https://linkinghub.elsevier.com/retrieve/pii/S1364661310000045}
	\BIBentrySTDinterwordspacing
	
	\bibitem[Kalman(1960)]{kalman_new_1960}
	\BIBentryALTinterwordspacing
	R.~E. Kalman, ``A {New} {Approach} to {Linear} {Filtering} and {Prediction}
	{Problems},'' \emph{Transactions of the ASME Journal of Basic Engineering},
	vol.~82, no. Series D, pp. 35--45, 1960. [Online]. Available:
	\url{http://fluidsengineering.asmedigitalcollection.asme.org/article.aspx?articleid=1430402}
	\BIBentrySTDinterwordspacing
	
	\bibitem[Kalman and Bucy(1961)]{kalman_new_1961}
	\BIBentryALTinterwordspacing
	R.~E. Kalman and R.~S. Bucy, ``\BIBforeignlanguage{en}{New {Results} in
		{Linear} {Filtering} and {Prediction} {Theory}},''
	\emph{\BIBforeignlanguage{en}{Journal of Basic Engineering}}, vol.~83, no.~1,
	pp. 95--108, Mar. 1961. [Online]. Available:
	\url{https://asmedigitalcollection.asme.org/fluidsengineering/article/83/1/95/426820/New-Results-in-Linear-Filtering-and-Prediction}
	\BIBentrySTDinterwordspacing
	
	\bibitem[Kurz et~al.(2016)Kurz, Gilitschenski, and
	Hanebeck]{kurz_recursive_2016}
	\BIBentryALTinterwordspacing
	G.~Kurz, I.~Gilitschenski, and U.~D. Hanebeck,
	``\BIBforeignlanguage{en}{Recursive {Bayesian} filtering in circular state
		spaces},'' \emph{\BIBforeignlanguage{en}{IEEE Aerospace and Electronic
			Systems Magazine}}, vol.~31, no.~3, pp. 70--87, Mar. 2016. [Online].
	Available: \url{https://ieeexplore.ieee.org/document/7475421}
	\BIBentrySTDinterwordspacing
	
	\bibitem[Azmani et~al.(2009)Azmani, Reboul, Choquel, and
	Benjelloun]{azmani_recursive_2009}
	M.~Azmani, S.~Reboul, J.~B. Choquel, and M.~Benjelloun, ``A recursive fusion
	filter for angular data,'' \emph{2009 IEEE International Conference on
		Robotics and Biomimetics, ROBIO 2009}, pp. 882--887, 2009, publisher: IEEE.
	
	\bibitem[Traa and Smaragdis(2013)]{traa_wrapped_2013}
	J.~Traa and P.~Smaragdis, ``A {Wrapped} {Kalman} {Filter} for {Azimuthal}
	{Speaker} {Tracking},'' \emph{IEEE Signal Processing Letters}, vol.~20,
	no.~12, pp. 1257--1260, 2013.
	
	\bibitem[Kurz et~al.(2014)Kurz, Gilitschenski, and
	Hanebeck]{kurz_recursive_2014}
	G.~Kurz, I.~Gilitschenski, and U.~D. Hanebeck, ``Recursive nonlinear filtering
	for angular data based on circular distributions,'' \emph{2013 American
		Control Conference}, pp. 5439--5445, 2014.
	
	\bibitem[Tronarp et~al.(2018)Tronarp, Hostettler, and
	Särkkä]{tronarp_continuous-discrete_2018}
	F.~Tronarp, R.~Hostettler, and S.~Särkkä, ``Continuous-{Discrete} von
	{Mises}-{Fisher} {Filtering} on {S2} for {Reference} {Vector} {Tracking},''
	\emph{2018 21st International Conference on Information Fusion, FUSION 2018},
	no. July, pp. 1345--1352, 2018.
	
	\bibitem[Hanzon and Hut(1991)]{hanzon_new_1991}
	\BIBentryALTinterwordspacing
	B.~Hanzon and R.~Hut, ``New {Results} on the {Projection} {Filter},''
	\emph{Research Memorandum}, vol. 105, no. March, pp. 9457--9475, 1991.
	[Online]. Available:
	\url{http://www.sciencedirect.com/science/article/pii/S0377221796004043}
	\BIBentrySTDinterwordspacing
	
	\bibitem[Brigo et~al.(1999)Brigo, Hanzon, and Le~Gland]{brigo_approximate_1999}
	\BIBentryALTinterwordspacing
	D.~Brigo, B.~Hanzon, and F.~Le~Gland, ``Approximate nonlinear filtering by
	projection on exponential manifolds of densities,'' \emph{Bernoulli}, vol.~5,
	no.~3, pp. 495--534, 1999. [Online]. Available:
	\url{http://projecteuclid.org/euclid.bj/1172617201}
	\BIBentrySTDinterwordspacing
	
	\bibitem[Tronarp and Särkkä(2020)]{tronarp_continuous-discrete_2020}
	\BIBentryALTinterwordspacing
	F.~Tronarp and S.~Särkkä, ``Continuous-{Discrete} {Filtering} and {Smoothing}
	on {Submanifolds} of {Euclidean} {Space},'' \emph{arXiv:2004.09335 [math,
		stat]}, Apr. 2020, arXiv: 2004.09335. [Online]. Available:
	\url{http://arxiv.org/abs/2004.09335}
	\BIBentrySTDinterwordspacing
	
	\bibitem[Nüsken et~al.(2019)Nüsken, Reich, and Rozdeba]{nusken_state_2019}
	\BIBentryALTinterwordspacing
	N.~Nüsken, S.~Reich, and P.~J. Rozdeba, ``\BIBforeignlanguage{en}{State and
		{Parameter} {Estimation} from {Observed} {Signal} {Increments}},''
	\emph{\BIBforeignlanguage{en}{Entropy}}, vol.~21, no.~5, p. 505, May 2019.
	[Online]. Available: \url{https://www.mdpi.com/1099-4300/21/5/505}
	\BIBentrySTDinterwordspacing
	
	\bibitem[Stratonovich(1960)]{stratonovich_conditional_1960}
	\BIBentryALTinterwordspacing
	R.~L. Stratonovich, ``\BIBforeignlanguage{en}{Conditional {Markov}
		{Processes}},'' \emph{\BIBforeignlanguage{en}{Theory of Probability \& Its
			Applications}}, vol.~5, no.~2, pp. 156--178, Jan. 1960. [Online]. Available:
	\url{http://epubs.siam.org/doi/10.1137/1105015}
	\BIBentrySTDinterwordspacing
	
	\bibitem[Kushner(1964)]{kushner_differential_1964}
	\BIBentryALTinterwordspacing
	H.~J. Kushner, ``\BIBforeignlanguage{en}{On the {Differential} {Equations}
		{Satisfied} by {Conditional} {Probablitity} {Densities} of {Markov}
		{Processes}, with {Applications}},'' \emph{\BIBforeignlanguage{en}{Journal of
			the Society for Industrial and Applied Mathematics Series A Control}},
	vol.~2, no.~1, pp. 106--119, Jan. 1964. [Online]. Available:
	\url{http://epubs.siam.org/doi/10.1137/0302009}
	\BIBentrySTDinterwordspacing
	
	\bibitem[Bain and Crisan(2009)]{bain_fundamentals_2009}
	A.~Bain and D.~Crisan, \emph{Fundamentals of stochastic filtering}, ser.
	Stochastic modelling and applied probability.\hskip 1em plus 0.5em minus
	0.4em\relax New York: Springer, 2009, no.~60, oCLC: ocn213479403.
	
	\bibitem[Brigo et~al.(1998)Brigo, Hanzon, and LeGland]{brigo_differential_1998}
	\BIBentryALTinterwordspacing
	D.~Brigo, B.~Hanzon, and F.~LeGland, ``A differential geometric approach to
	nonlinear filtering: the projection filter,'' \emph{IEEE Transactions on
		Automatic Control}, vol.~43, no.~2, pp. 247--252, 1998. [Online]. Available:
	\url{http://ieeexplore.ieee.org/document/661075/}
	\BIBentrySTDinterwordspacing
	
	\bibitem[van Handel and Mabuchi(2005)]{van_handel_quantum_2005}
	\BIBentryALTinterwordspacing
	R.~van Handel and H.~Mabuchi, ``\BIBforeignlanguage{en}{Quantum projection
		filter for a highly nonlinear model in cavity {QED}},''
	\emph{\BIBforeignlanguage{en}{Journal of Optics B: Quantum and Semiclassical
			Optics}}, vol.~7, no.~10, pp. S226--S236, Oct. 2005. [Online]. Available:
	\url{https://iopscience.iop.org/article/10.1088/1464-4266/7/10/005}
	\BIBentrySTDinterwordspacing
	
	\bibitem[Amari(1990)]{amari_differential-geometrical_1990}
	S.~Amari, \emph{Differential-geometrical methods in statistics}, 2nd~ed., ser.
	Lecture notes in statistics.\hskip 1em plus 0.5em minus 0.4em\relax Berlin ;
	New York: Springer-Verlag, 1990, no.~28.
	
	\bibitem[Wong and Zakai(1965)]{wong_relation_1965}
	\BIBentryALTinterwordspacing
	E.~Wong and M.~Zakai, ``\BIBforeignlanguage{en}{On the relation between
		ordinary and stochastic differential equations},''
	\emph{\BIBforeignlanguage{en}{International Journal of Engineering Science}},
	vol.~3, no.~2, pp. 213--229, Jul. 1965. [Online]. Available:
	\url{https://linkinghub.elsevier.com/retrieve/pii/0020722565900455}
	\BIBentrySTDinterwordspacing
	
	\bibitem[Gatto and Jammalamadaka(2007)]{gatto_generalized_2007}
	\BIBentryALTinterwordspacing
	R.~Gatto and S.~R. Jammalamadaka, ``\BIBforeignlanguage{en}{The generalized von
		{Mises} distribution},'' \emph{\BIBforeignlanguage{en}{Statistical
			Methodology}}, vol.~4, no.~3, pp. 341--353, Jul. 2007. [Online]. Available:
	\url{https://escholarship.org/uc/item/8m91b0p7}
	\BIBentrySTDinterwordspacing
	
	\bibitem[Gatto(2008)]{gatto_computational_2008}
	R.~Gatto, ``\BIBforeignlanguage{en}{Some computational aspects of the
		generalized von {Mises} distribution},''
	\emph{\BIBforeignlanguage{en}{Statistics and Computing}}, vol.~18, no.~3, pp.
	321--331, Sep. 2008.
	
	\bibitem[Kloeden and Platen(2010)]{kloedenNumericalSolutionStochastic2010}
	P.~E. Kloeden and E.~Platen, \emph{\BIBforeignlanguage{eng}{Numerical solution
			of stochastic differential equations}}, corr. 3. print~ed., ser. Applications
	of mathematics.\hskip 1em plus 0.5em minus 0.4em\relax Berlin: Springer,
	2010, no.~23.
	
	\bibitem[Kutschireiter et~al.(2021)Kutschireiter, Basnak, Wilson, and
	Drugowitsch]{kutschireiterBayesianPerspectiveRing2021}
	\BIBentryALTinterwordspacing
	A.~Kutschireiter, M.~A. Basnak, R.~I. Wilson, and J.~Drugowitsch,
	``\BIBforeignlanguage{en}{A {Bayesian} perspective on the ring attractor for
		heading-direction tracking in the {Drosophila} central complex},'' Tech.
	Rep., Dec. 2021, \emph{bioRxiv}, Cold Spring Harbor Laboratory, Section: New
	Results Type: article. [Online]. Available:
	\url{https://www.biorxiv.org/content/10.1101/2021.12.17.473253v1}
	\BIBentrySTDinterwordspacing
	
	\bibitem[Doucet et~al.(2010)Doucet, Godsill, and
	Andrieu]{doucet_sequential_2010}
	A.~Doucet, S.~Godsill, and C.~Andrieu, ``\BIBforeignlanguage{en}{On sequential
		{Monte} {Carlo} sampling methods for {Bayesian} filtering},''
	\emph{\BIBforeignlanguage{en}{Statistics and Computing}}, p.~12, 2010.
	
\end{thebibliography}


\end{document}